%% file: neurips_2026.tex
\documentclass{article}

 \usepackage[preprint]{neurips_2026}

\usepackage[utf8]{inputenc} 
\usepackage[T1]{fontenc}    
\usepackage{hyperref}       
\usepackage{url}            
\usepackage{booktabs}       
\usepackage{amsfonts}       
\usepackage{nicefrac}       
\usepackage{microtype}      
\usepackage{xcolor}         
\usepackage{array}
\usepackage{multirow}
\usepackage[table]{xcolor}
\usepackage{pifont}
\usepackage{colortbl}
\usepackage{graphicx}
\usepackage{caption}
\usepackage{makecell}
\usepackage{tabularx}
\usepackage{amsmath}
\usepackage{threeparttable}
\usepackage{algorithm}
\usepackage{algpseudocode}

\definecolor{jptagreen}{HTML}{00796B}
\definecolor{jptaorange}{HTML}{D86A4A}
\definecolor{jptagray}{HTML}{637085}
\definecolor{jptablue}{HTML}{2F77A5}
\definecolor{jptadark}{HTML}{252D3A}

\definecolor{JPTADrop}{HTML}{B00020}
\definecolor{JPTAGain}{HTML}{0057B8}
\newcommand{\nd}[1]{\textcolor{JPTADrop}{#1}}
\newcommand{\pd}[1]{\textcolor{JPTAGain}{+#1}}

\definecolor{oursbg}{RGB}{237,244,250}
\definecolor{avgbg}{RGB}{247,240,226}
\definecolor{posgreen}{RGB}{0,120,40}
\definecolor{negred}{RGB}{180,30,30}

\newcommand{\fsBest}[1]{\textbf{#1}}
\newcommand{\fsSecond}[1]{\underline{#1}}

\newcommand{\fsAvg}[1]{\cellcolor{avgbg}#1}

\newcommand{\fsAvgSecond}[1]{\cellcolor{avgbg}\fsSecond{#1}}

\newcommand{\fsJPos}[2]{%
\begin{tabular}[c]{@{}c@{}}
\fsBest{#1}\\[-0.45ex]
{\tiny\textcolor{posgreen}{$\Delta{+}#2$}}
\end{tabular}%
}

\newcommand{\fsJNegSecond}[2]{%
\begin{tabular}[c]{@{}c@{}}
\fsSecond{#1}\\[-0.45ex]
{\tiny\textcolor{negred}{$\Delta{-}#2$}}
\end{tabular}%
}

\newcommand{\fsJNeg}[2]{%
\begin{tabular}[c]{@{}c@{}}
#1\\[-0.45ex]
{\tiny\textcolor{negred}{$\Delta{-}#2$}}
\end{tabular}%
}

\newcommand{\fsAvgJPos}[2]{\cellcolor{avgbg}\fsJPos{#1}{#2}}

\newcommand{\cmark}{\ding{51}}%
\newcommand{\xmark}{\ding{55}}%
\title{Batch-Conditioned Semantic Anchors for Robust Transductive Adaptation of Vision--Language Models}

\author{%
  Mohammed Rahman Sherif Khan Mohammad \\
  Edge Hill University, UK \\
  \texttt{Mohamm@edgehill.ac.uk}
  \And
  Ardhendu Behera \thanks{Corresponding author.}\\   
  Edge Hill University, UK \\
  \texttt{Beheraa@edgehill.ac.uk}
  \AND
  Sandip Pradhan \\
  Edge Hill University, UK \\
  \texttt{Pradhans@edgehill.ac.uk}
  \And
  Swagat Kumar \\
  Edge Hill University, UK \\
  \texttt{Kumars@edgehill.ac.uk}
  \And
  Amr Ahmed \\
  Edge Hill University, UK \\
  \texttt{Ahmeda@edgehill.ac.uk}
}

\begin{document}

\maketitle

\input{sec/0_abstract}
\input{sec/1_intro}
\input{sec/2_related_works}
\input{sec/3_method}
\input{sec/4_experiments}

\input{sec/5_conclusion}

\begin{ack}
\vspace{-0.5em}
This work was supported by UK Research and Innovation (UKRI) through
the Engineering and Physical Sciences Research Council (EPSRC) grant
ATRACT (\texttt{EP/X028631/1}), the Economic and Social Research Council (ESRC) grant SCAnDi (\texttt{ES/Y010655/1}), and Cancer Research UK (CRUK) grants \texttt{EDDISA-May21\textbackslash100001} and
\texttt{EDDPMA-May22\textbackslash100028}.
\end{ack}

{
\small
\bibliographystyle{plainnat}
\bibliography{main}
}
\clearpage


\appendix

\input{sec/6_appendix}



\end{document}

%% file: sec/0_abstract.tex
\begin{abstract}

Transductive adaptation improves vision–language models at test time by refining predictions on an unlabeled batch, but current methods face a key trade-off. Aggressive approaches exploit batch structure but can drift when batches are sparse, have prior shift, or cover little of the label space. Conservative methods better preserve CLIP text semantics, but fixed semantic references can under-adapt even when the batch provides strong corrective evidence.
We propose \textbf{JPTA} (\emph{Joint Prior and Transport Adaptation}), a prior-aware framework that casts transductive VLM adaptation as \emph{batch-conditioned semantic reference estimation}. Instead of treating CLIP text prototypes as fixed priors or freely replaceable, JPTA views them as semantic references that shift toward image-side batch structure only under calibrated evidence. It estimates batch priors and soft image-side prototypes to form transported semantic references whose impact is scaled by batch reliability. 
Across standard benchmarks, low effective-class regimes, all-class evaluation, and online streams, JPTA consistently outperforms TransCLIP and StatA, with the largest gains on sparse and prior-shifted batches. Anchor-drift, active-support, and absent-class-mass diagnostics show that JPTA fixes effective-class mismatch without unconstrained transductive drift, indicating that robust transductive VLM adaptation should recalibrate semantic references using reliable batch evidence rather than choose between unrestricted transduction and fixed prototypes. Code is available at \url{https://github.com/MR-Sherif/JPTA.git}.
\end{abstract}
%
%

%% file: sec/1_intro.tex
\section{Introduction}
\label{sec:intro}
\vspace{-1em}

\begin{figure}[t]
    \centering
    \includegraphics[width=\linewidth]{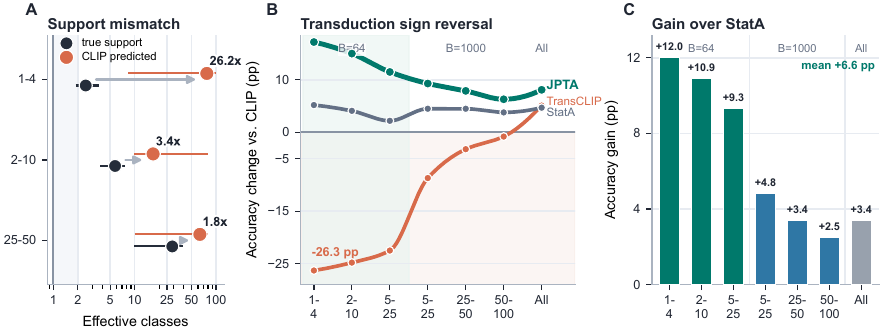}
    \caption{
        \textbf{Low-coverage batches cause class mismatch and semantic drift; JPTA fixes both.}
        \textbf{A:} In sparse batches, the true effective class support (black) is much smaller than that estimated by zero-shot CLIP predictions (orange), revealing strong class-support inflation; labels show the median inflation factor.
        \textbf{B:} This mismatch alters transductive adaptation: TransCLIP~\cite{zanella2024boosting} (orange) hurts performance in low-coverage regimes but recovers as coverage grows, whereas JPTA (green) remains consistently beneficial.
        \textbf{C:} JPTA outperforms StatA~\cite{zanella2025realistic} in both sparse-batch and all-class regimes, with results grouped by $B=64$ (green), $B=1000$ (blue), and all-class (gray).
    }
    \label{fig:jpta_first_page}
    \vspace{-2.0em}
\end{figure}

Vision--language models (VLMs) such as CLIP have transformed visual recognition by transferring natural-language supervision to diverse downstream tasks \cite{radford2021learningclip}. Many methods adapt VLMs with prompt learning, cache models, and lightweight adapters \cite{zhou2022learningcoop,zhou2022conditionalCoCoOp,zhang2022tip,song2023meta}, but typically assume an \emph{inductive} setting with labeled support examples. In practice, labels are often unavailable at test time: the model receives only unlabeled batches from unknown environments, with domain shift, class imbalance, and varying batch composition. This motivates \emph{transductive} adaptation, where predictions are refined jointly over the current unlabeled batch rather than per sample.

Transductive inference is attractive because the test batch reveals useful structure: nearby samples in the VLM feature space should share predictions, and the batch can indicate which classes are present. This idea has a long history in few-shot learning via graph propagation, manifold regularization, prototype refinement, and information-theoretic objectives \cite{satorras2018few,rodriguez2020embedding,ziko2020laplacian,boudiaf2020information,kang2021relational,zhu2023transductive}. Recently, it has been applied to CLIP through transductive zero-shot and few-shot inference \cite{martin2024transductive}, realistic test-time adaptation \cite{zanella2025realistic}, and language-aware transductive objectives \cite{baklouti2026limo}. These works show that unlabeled test batches can strongly guide adaptation even without labeled support or gradient-based fine-tuning.

Despite this promise, transductive VLM adaptation remains brittle in realistic settings. Batch usefulness depends heavily on its composition: large, representative batches yield reliable image-side geometry, while batches with few labels, class-prior shift, or noisy neighborhoods make transductive dynamics misleading. In such \emph{low effective-class} regimes, locally coherent but semantically incorrect groups can dominate refinement, pushing assignments and class statistics away from the VLM’s pretrained language-aligned structure \cite{veilleux2021realistic,zanella2025realistic}.

Existing methods face a sharp trade-off between \emph{adaptability} and \emph{stability}. Aggressive methods like TransCLIP \cite{zanella2024boosting} jointly refine assignments, class statistics, and neighborhood consistency to exploit batch structure, but can drift when batches are sparse, imbalanced, or semantically misleading. Conservative methods like StatA \cite{zanella2025realistic} gain robustness by anchoring refinement to the original CLIP text semantics, but this anchor is fixed. We identify the core failure as \emph{semantic-reference mismatch}: low-coverage batches inflate the predicted class support, reverse the benefit of aggressive transduction, and create the largest gap between fixed anchoring and JPTA.
In low-coverage batches, image-side geometry can reveal which classes are active, yet unconstrained transduction turns this signal into semantic drift. Conversely, fixed text anchoring prevents drift but also prevents the semantic reference from correcting systematic zero-shot mismatch. Thus, the key object to adapt is not only the assignment matrix or class statistics, but the semantic reference itself.

We propose \textbf{JPTA} (\emph{Joint Prior and Transport Adaptation}), a transductive VLM adaptation framework based on batch-conditioned semantic anchoring. JPTA keeps the usual transductive inference setting but changes what is adapted: instead of fixing CLIP text prototypes, it estimates a batch-conditioned semantic anchor via reliability-controlled transport. Concretely, JPTA estimates the batch class prior, builds soft image-side class prototypes from evolving assignments, and transports text prototypes toward the batch geometry only when the evidence is reliable. The transported anchor is then shrunk toward the original CLIP geometry and used in a prior-aware refinement that couples semantic anchoring, graph consistency, class likelihood, and batch-marginal regularization. This lets the model stay conservative on unreliable batches while adapting when batches consistently support a corrected semantic configuration.

\noindent\textbf{Contributions.}
Our contributions are threefold. First, we identify \emph{semantic-reference mismatch} as a key failure mode of transductive VLM adaptation under sparse, prior-shifted batches: free transduction drifts toward hallucinated class support, while fixed anchoring cannot repair misaligned zero-shot references. Second, we introduce \emph{batch-conditioned semantic anchoring}, a principle that treats CLIP text prototypes as movable yet constrained semantic references, implemented via reliability-weighted orthogonal transport and shrinkage toward the original CLIP text geometry. Third, we design a prior-aware transductive solver that couples transported anchors, graph consistency, class-likelihood refinement, and batch-marginal regularization. Across sparse, high-coverage, all-class, and online regimes, JPTA outperforms fixed-anchor and free-transduction baselines, with diagnostics showing reduced absent-class mass and controlled semantic drift.

%% file: sec/2_related_works.tex
\vspace{-0.5em}
\section{Related Work}
\label{sec:related_work}
\vspace{-0.5em}
\textbf{Inductive few-shot adaptation of vision--language models.}
Following CLIP \cite{radford2021learningclip}, many works study adapting VLMs with limited labeled data. Prompt-learning methods such as CoOp \cite{zhou2022learningcoop}, CoCoOp \cite{zhou2022conditionalCoCoOp}, MaPLe \cite{khattak2023maple}, and self-regularized or consistency-guided prompting \cite{khattak2023self,roy2024coprompt} improve zero-shot transfer by learning task-specific contexts around frozen text and image encoders. Complementary work proposes lightweight feature adaptation and cache-based refinement, including CLIP-Adapter \cite{gao2024clip-adapter}, TIP-Adapter \cite{zhang2022tip}, Meta-Adapter \cite{song2023meta}, TaskRes \cite{yu2023task}, low-rank adaptation \cite{zanella2024lowclip-lora}, graph-based tuning \cite{li2023graphadapter}, training-only graph supervision \cite{Mohammad_2026_CVPR}, and dual-memory mechanisms \cite{zhang2024dual}. Recent studies also question evaluation practices and hidden design choices in few-shot VLM adaptation \cite{silva2024closerCLAP,kravets2025rethinking,farina2025rethinking}. Our setting is fundamentally different: these methods mainly address the \emph{inductive} regime with labeled support sets, whereas we study \emph{transductive} adaptation on an unlabeled test batch.

\textbf{Transductive prediction and batch-wise inference.}
Transductive inference has a long history in few-shot learning, where unlabeled query batches are exploited via graph propagation, manifold smoothing, iterative prototype correction, and information-theoretic objectives \cite{satorras2018few,rodriguez2020embedding,ziko2020laplacian,boudiaf2020information,liu2020prototype,hu2021leveraging,kang2021relational,zhu2023transductive,lazarou2021iterative}. This work established that query samples should be inferred jointly whenever batch geometry carries useful structure, and that realistic class composition is crucial, since transductive gains depend strongly on how query batches are formed \cite{veilleux2021realistic}. This principle was extended to CLIP in Transductive Zero-Shot and Few-Shot \cite{martin2024transductive} and strengthened by LIMO \cite{baklouti2026limo}, which showed that language-aware transductive objectives improve few-shot CLIP adaptation. Our work adopts this transductive view but targets a fully unlabeled test-time setting, where the batch alone must guide inference.

\textbf{Test-time adaptation of vision--language models.}
Recent work extends adaptation from few-shot, support-based settings to unlabeled test time. One direction is \emph{test-time prompt tuning}, which adapts textual context at inference to improve zero-shot robustness, including TPT \cite{shu2022test} and its calibration-focused variants \cite{yoon2024ctpt,sharifdeen2025otpt,ahamed2026atpt,sheng2025rtpt}. Another direction performs single-image or instance-level adaptation in feature or parameter space, such as CLIP Reward \cite{zhao2024clipreward}, UMFC \cite{liang2024umfc}, WATT \cite{osowiechi2024watt}, CLIPTTA \cite{lafon2025cliptta}, and ROSITA \cite{sreenivas2025rosita}. These methods improve robustness to domain shift, calibration error, or open-set conditions but do not explicitly treat the problem as batch-wise transductive inference over a current unlabeled set. By contrast, SCAP \cite{zhang2025scap} and TransCLIP \cite{zanella2024boosting} directly exploit batch structure, showing that neighborhood agreement and joint refinement can significantly outperform per-sample adaptation when the batch is informative. This distinction is central for our setting: JPTA is a \emph{batch-transductive} VLM adaptation method, rather than a generic per-sample test-time adaptation approach.

\textbf{Robust transductive adaptation under realistic batch regimes.}
The closest prior work is StatA \cite{zanella2025realistic}, which shows that aggressive transduction is brittle when the batch has few effective classes, a strong prior shift, or a correlated structure. StatA stabilizes adaptation by anchoring transductive refinement to the original text-side CLIP semantics, improving robustness over free transduction, but its fixed anchor cannot adjust when the batch offers reliable evidence that the semantic reference should shift. Our method differs from both TransCLIP \cite{zanella2024boosting} and StatA \cite{zanella2025realistic} by making the anchor itself \emph{batch-aware}: JPTA estimates the batch prior, transports text prototypes toward the image-side batch structure, and uses this transported anchor in prior-aware transductive refinement. This situates JPTA between free transduction and rigid anchoring, retaining the robustness of anchoring while regaining the flexibility to exploit reliable batch evidence.

\textbf{Comparison scope.} We focus on methods with the same access pattern: frozen CLIP features, no labeled support set, and batch-wise transductive inference. Under this setting, TransCLIP and StatA are the closest baselines. Broader test-time adaptation methods (TPT, WATT, CLIPTTA, and related prompt/feature adaptation approaches) address a different problem because they update prompts, features, or model parameters, or adapt per image instead of using a batch-level transductive solver. We still report compatible representatives with an access-pattern audit in the Appendix (Table~\ref{tab:baseline_access_audit_full}). 

%% file: sec/3_method.tex
\vspace{-0.75em}
\section{Batch-Conditioned Semantic Anchoring}
\label{sec:method}
\vspace{-0.75em}
Given an unlabeled test batch and a fixed CLIP label space, JPTA estimates a batch-conditioned semantic reference. It starts from zero-shot predictions to estimate the batch prior, builds soft image prototypes, and transports CLIP text prototypes toward reliable batch geometry with trust-controlled shrinkage. These anchors then guide prior-aware assignment refinement and adaptive class-statistic updates, avoiding both unconstrained transductive drift and rigid anchoring.
\vspace{-0.3em}
\subsection{Batch-transductive setting}
\vspace{-0.3em}
We adopt the transductive VLM setting of TransCLIP and StatA \cite{zanella2024boosting,zanella2025realistic}. Let $f(\cdot)$ and $g(\cdot)$ be the frozen image and text encoders of a pretrained VLM. For label set $\mathcal{Y}=\{1,\dots,K\}$, we encode one normalized text prototype per class, $T=[t_1,\dots,t_K]\in\mathbb{R}^{d\times K}$. At test time, we observe an unlabeled batch of normalized image features
\(
\mathcal{B}=\{x_i\}_{i=1}^N,\ x_i=f(I_i)\in\mathbb{R}^d,
\)
with no labeled support set. The goal is to jointly infer labels for all samples in $\mathcal{B}$ rather than independently.

As in prior work, the full label space may be large, while a given batch may contain only a small subset of active classes. We call this a \emph{low effective-class regime}, where the batch prior can differ substantially from the global label prior \cite{zanella2025realistic}. In this regime, free transduction may drift, and fixed anchors may become overly conservative. 
We initialize inference from the zero-shot VLM predictions
\begin{equation}
\ell^{(0)}_{ik}=\tau x_i^\top t_k,\qquad
A^{(0)}_{ik}=\frac{\exp(\ell^{(0)}_{ik})}{\sum_{c=1}^K \exp(\ell^{(0)}_{ic})},
\label{eq:zeroshot_anchor}
\end{equation}
where $\tau$ is the logit scale and $A^{(0)}\in\mathbb{R}^{N\times K}$ is the zero-shot anchor. We build a batch $k$-nearest-neighbor graph with affinity matrix $W\in\mathbb{R}^{N\times N}$ and soft assignment matrix $Z\in\mathbb{R}^{N\times K}$ whose rows lie on the simplex. This transductive scaffold fixes the inference protocol but leaves one key choice: \emph{what should adapt at test time?} Existing solvers mainly update assignments and class statistics, either from image-side geometry \cite{zanella2024boosting} or under a fixed text reference \cite{zanella2025realistic}. JPTA instead adapts the semantic reference itself: CLIP text prototypes become batch-conditioned anchors estimated via reliability-controlled transport, and assignments are refined relative to these calibrated references.
\vspace{-0.3em}
\subsection{Regime statistics and batch prior}
\vspace{-0.3em}
The current batch provides more than local geometry: it shows how many classes appear active and how reliable the transductive signal is. From the anchor probabilities $A^{(0)}$, we compute hard pseudo-labels
\(
\hat y_i=\arg\max_k A^{(0)}_{ik},
\)
and estimate several regime statistics. We first measure the \emph{neighbor agreement} $\rho$ on the batch graph, then compute the \emph{anchor ambiguity} $\bar m$ as the mean margin between the top-1 and top-2 probabilities.
\begin{equation}
\rho = \frac{1}{|\mathcal{E}|}\sum_{(i,j)\in\mathcal{E}} \mathbf{1}[\hat y_i=\hat y_j],\ 
\bar m = \frac{1}{N}\sum_{i=1}^N \left(A^{(0)}_{i,(1)}-A^{(0)}_{i,(2)}\right),
\label{eq:margin}
\end{equation}
where $A^{(0)}_{i,(1)}$ and $A^{(0)}_{i,(2)}$ are the largest and second-largest entries of row $i$. Third, we estimate the effective class coverage of the current batch through the number of predicted classes, maintaining a soft batch prior by averaging the current class posteriors initialized with $Z=A^{(0)}$.
\begin{equation}
K_{\mathrm{pred}} = \left|\{\hat y_i : i=1,\dots,N\}\right|,
\quad
C_{\mathrm{hard}} = \frac{K_{\mathrm{pred}}}{K}, \quad 
\pi_k = \frac{\frac{1}{N}\sum_{i=1}^N z_{ik}+\alpha_\pi/K}{{1+\alpha_\pi}}, 
\label{eq:hard_coverage}
\end{equation}
where $\alpha_\pi$ is a fixed symmetric prior strength. $C_{\mathrm{hard}}$ indicates whether the batch covers most of the label space. High hard coverage triggers conservative fallback and adaptive iteration control, while low hard coverage uses the full refinement budget, since sparse batches may still contain reliable evidence for their active classes. From $\pi_k$, we derive a soft effective-class estimate $C_\mathrm{soft}$ to compute the transport trust coefficient:
\begin{equation}
\eta =1-\frac{C_{\mathrm{soft}}}{K}, \qquad C_{\mathrm{soft}} = \exp(H(\pi)),
\qquad
H(\pi)=-\sum_{k=1}^K \pi_k \log \pi_k.
\label{eq:soft_coverage}
\end{equation}

Thus, \(\rho\), $\bar m$, and samples per predicted class \(N/K_\mathrm{pred}\) control \emph{whether and how aggressively to refine}, \(C_{\mathrm{hard}}\) controls the \emph{batch regime}, and \(C_{\mathrm{soft}}\) controls \emph{how much the semantic anchor is allowed to move}. In the next section, $\eta$ is employed to calculate the transported semantic anchors. 

\vspace{-0.3em}
\subsection{Transported semantic anchors}
\vspace{-0.3em}
The core idea of JPTA is to adapt the semantic anchor itself. Given current assignments $Z$, we form soft image-side class prototypes by averaging the image features weighted by their class posteriors: 
\(
\bar x_k = \frac{\sum_{i=1}^N z_{ik}x_i}{\sum_{i=1}^N z_{ik}+\varepsilon},
\
n_k = \sum_{i=1}^N z_{ik},
\)
where $n_k$ is the effective support of class $k$. 
We seek an orthogonal transport that aligns the original text prototypes with the image-side prototypes while preserving the text space geometry. We solve the weighted Procrustes problem
\begin{equation}
R^\star=\arg\min_{R\in\mathcal{O}(d)} \sum_{k=1}^K w_k \|Rt_k-\bar x_k\|_2^2, \quad \hat t_k = R^\star t_k, \quad \tilde t_k = \mathrm{norm}\!\left((1-\eta_k)t_k+\eta_k\hat t_k\right),
\label{eq:procrustes}
\end{equation}
where $Rt_k$ is the transported text prototype of class $k$, obtained by applying the global orthogonal alignment $R$ to $t_k$. Here $\mathcal{O}(d)$ is the orthogonal group and $w_k$ are confidence weights based on class support and batch reliability. Let
\(
M=\sum_{k=1}^K w_k \bar x_k t_k^\top 
\),
and if $M=U\Sigma V^\top$ is the singular value decomposition, then the optimal transport is $R^\star=UV^\top$.
The transported text prototypes are $\hat t_k$.
To reduce sensitivity to unreliable batches, we shrink the transported prototypes toward the original text prototypes $t_k$. 
$\eta_k\in[0,1]$ is a transport-trust coefficient: it increases when class $k$ is well supported and the batch regime is reliable, and decreases when the batch is ambiguous or poorly covered. This yields a \emph{transported semantic anchor}: unlike StatA \cite{zanella2025realistic}, which keeps $t_k$ fixed, JPTA lets the anchor move toward the image-side batch geometry while still shrinking toward the original CLIP semantics.

Using $\tilde T=[\tilde t_1,\dots,\tilde t_K]$, we obtain transported anchor probabilities $A^{(\mathrm{tr})}_{ik}$. Instead of discarding the original zero-shot anchor, we combine both probability sets geometrically in log space,
\begin{equation}
A^{(\mathrm{tr})}_{ik}
=
\frac{\exp(\tau x_i^\top \tilde t_k)}
{\sum_{c=1}^K \exp(\tau x_i^\top \tilde t_c)}, \quad
\log A^{(\mathrm{anc})}_{ik}
\propto
(1-\eta)\log A^{(0)}_{ik}
+
\eta \log A^{(\mathrm{tr})}_{ik},
\label{eq:transport_anchor}
\end{equation}
where $\eta$ is a batch-level trust coefficient controlled by the regime statistics. Eq.~\eqref{eq:transport_anchor} is central to JPTA: the anchor is no longer fixed, but it is also not free to drift arbitrarily.

\textbf{Why shrinkage controls semantic drift?}
The transported prototype can improve semantic reference when the batch estimate is reliable, but should not move freely. The following analysis formalizes this reliability–drift trade-off.

\textbf{Proposition 1.} Let \(t_k,\hat t_k,u_k\in\mathbb{S}^{d-1}\) denote the original CLIP text prototype, the transported prototype, and the unknown image-side semantic center for class \(k\). Define
\[
\tilde t_k(\eta_k)=
\mathrm{norm}\big((1-\eta_k)t_k+\eta_k\hat t_k\big),
\qquad \eta_k\in[0,1],
\]
and let
\[
\alpha_k=\angle(t_k,u_k),\quad
\beta_k=\angle(\hat t_k,u_k),\quad
\theta_k=\angle(t_k,\hat t_k).
\]
If the transported estimate is better aligned with the image-side centre,
\(\beta_k<\alpha_k<\pi/2\), then there exists \(\eta_k>0\) such that
\(
\angle(\tilde t_k(\eta_k),u_k)<\angle(t_k,u_k).
\)
Moreover, the semantic drift from the original CLIP anchor is exactly
\[
\delta_k(\eta_k)
=
\angle(\tilde t_k(\eta_k),t_k)
=
\arctan
\frac{\eta_k\sin\theta_k}
{1-\eta_k+\eta_k\cos\theta_k},
\]
equivalently,
\[
\delta_k(\eta_k)
=
\arccos
\frac{1-\eta_k+\eta_k\cos\theta_k}
{\sqrt{(1-\eta_k)^2+\eta_k^2+2\eta_k(1-\eta_k)\cos\theta_k}} .
\]
Thus, shrinkage deterministically bounds semantic movement: \(\delta_k(\eta_k)=O(\eta_k\sin\theta_k)\) for small \(\eta_k\). Reducing \(\eta_k\) under weak support or poor alignment limits worst-case anchor drift (proof in Appx~\ref{app:shrinkage_drift}).
\vspace{-0.3em}
\subsection{Prior-aware transductive refinement}
\label{sec:prior_refinement}
\vspace{-0.3em}
We refine the assignment matrix \(Z\in\Delta_K^N\) using four evidence sources: class likelihood, graph consistency, semantic anchoring, and batch-prior regularization. Class-conditional image features follow diagonal Gaussians
\(
q(x_i\mid k)=\mathcal{N}(x_i;\mu_k,\Sigma_k),
\)
with \((\mu_k,\Sigma_k)\) updated from current soft assignments. We take \(\ell_{ik}\) from Eq.~\eqref{eq:zeroshot_anchor}, the anchor distribution \(A^{(\mathrm{anc})}_{ik}\) from Eq.~\eqref{eq:transport_anchor}, and let \(L_G\) be the graph Laplacian of affinity matrix \(W\). For fixed anchor and prior, JPTA solves
\begin{equation}
\min_{Z\in\Delta_K^N}
-\left\langle Z,\ell+\lambda_a \log A^{(\mathrm{anc})}\right\rangle
+
\lambda_g\,\mathrm{Tr}(Z^\top L_G Z)
+
\lambda_\pi\,\mathrm{KL}\!\left(
\frac{1}{N}Z^\top\mathbf{1}\,\middle\|\,\pi
\right),
\label{eq:jpta_objective}
\end{equation}
where \(\lambda_a\), \(\lambda_g\), and \(\lambda_\pi\) control the strength of semantic anchoring, graph smoothness, and batch-marginal consistency, respectively. 
We optimize Eq.~\eqref{eq:jpta_objective} by alternating assignment refinement with updates of the class statistics and prior. At iteration \(t\), we fix the current likelihood \(\ell^{(t)}\), anchor \(A_{\mathrm{anc}}^{(t)}\), prior \(\pi^{(t)}\), and graph message
\(
G^{(t)}=-L_G Z^{(t)} .
\)
We then apply an exponentiated-gradient update
\begin{equation}
\tilde{Z}_{ik}^{(t+1)}
=
\exp\left(
\ell_{ik}^{(t)}
+\lambda_a \log A_{\mathrm{anc},ik}^{(t)}
+\lambda_g G_{ik}^{(t)}
-\lambda_\pi
\log
\frac{\bar z_k^{(t)}+\epsilon}{\pi_k^{(t)}+\epsilon}
\right),
\label{eq:jpta_update}
\end{equation}
where \(\bar z_k^{(t)}=\frac{1}{N}\sum_i z_{ik}^{(t)}\). We normalize the rows of \(\tilde Z^{(t+1)}\) to obtain \(Z^{(t+1)}\), and absorb constant graph and KL factors into \(\lambda_g\) and \(\lambda_\pi\). Eq.~\eqref{eq:jpta_objective} highlights each term: likelihood updates class statistics, graph smoothing enforces local consistency, semantic anchoring links assignments to the batch-conditioned reference, and prior matching prevents drift toward hallucinated classes. We adapt the prior: in reliable regimes we use Sinkhorn-style normalization for stronger marginal matching; in weak or ambiguous regimes we keep a softer KL penalty.
\vspace{-0.3em}
\subsection{Adaptive anchored class statistics}
\vspace{-0.3em}
Given updated assignments $Z$, we compute the empirical batch moments,
\begin{equation}
\bar\mu_k = \frac{\sum_i z_{ik}x_i}{n_k+\varepsilon},
\qquad
\bar\Sigma_k = \frac{\sum_i z_{ik}(x_i-\bar\mu_k)^{\odot 2}}{n_k+\varepsilon},
\label{eq:empirical_moments}
\end{equation}
where $\odot 2$ denotes element-wise squaring. Following StatA’s statistical anchoring principle, we update class moments via shrinkage instead of direct replacement, anchoring to the \emph{transported} prototypes $\tilde t_k$ rather than the original text prototypes. Let $\alpha_k>0$ be an adaptive pseudocount and
\(
\beta_k = \frac{n_k}{n_k+\alpha_k}.
\)
The anchored updates are
\begin{equation}
\mu_k \leftarrow \mathrm{norm}\!\left(\beta_k \bar\mu_k + (1-\beta_k)\tilde t_k\right), \qquad
\Sigma_k \leftarrow
\beta_k \bar\Sigma_k
+
(1-\beta_k)\Sigma_k^{(0)}
+
\beta_k(1-\beta_k)(\bar\mu_k-\tilde t_k)^{\odot 2},
\label{eq:anchored_updates}
\end{equation}
where $\Sigma_k^{(0)}$ is the initial covariance anchor. The pseudocount $\alpha_k$ increases in ambiguous or low-coverage settings and decreases when the batch provides stronger support. JPTA thus ranges from conservative anchored adaptation to flexible batch-driven refinement.

\textbf{Relation to previous work. }
JPTA strictly generalizes the TransCLIP--StatA continuum \cite{zanella2024boosting, zanella2025realistic}. With zero transport trust and a weakened prior, it behaves like an anchored solver similar to StatA; removing anchor shrinkage and letting the transported anchor dominate yields a more aggressive transductive regime. The contribution is not separate controls but a single organizing principle: \emph{batch-conditioned semantic anchoring}. Regime statistics only determine when the batch adjusts the anchor and how strongly the prior guides inference.

%% file: sec/4_experiments.tex
\vspace{-0.5em}
\section{Evaluation Under Realistic Batch Regimes}
\label{sec:experiments}
\vspace{-0.5em}

We follow the transductive VLM protocol of TransCLIP~\cite{zanella2024boosting} and StatA~\cite{zanella2025realistic}; unless otherwise stated, all main experiments use the same frozen CLIP ViT-B/16 backbone, text prompts, precomputed image features, datasets, and batch-generation procedure as the baselines. We evaluate on ImageNet and ten transfer benchmarks (listed in the Appendix). 
Baselines include shared non-adaptive references, CLIP \cite{radford2021learningclip} and MTA \cite{zanella2024testMTA}; frozen-feature transductive methods, Dirichlet \cite{martin2024transductive}, ZLaP \cite{kalantidis2024labelZLap}, TransCLIP \cite{zanella2024boosting}, and StatA \cite{zanella2025realistic}; and online stream methods, TENT \cite{wang2020tent}, TDA \cite{karmanov2024efficient}, and DMN \cite{zhang2024dual}. We focus on TransCLIP and StatA in the main text as the most comparable VLM-specific batch-transductive baselines under the same frozen-feature, no-support-label protocol. We report top-1 accuracy for independent batch adaptation, low effective-class batches, and realistic online streams. For batch evaluation, we vary batch size $B$ and effective active classes $K_{\mathrm{eff}}$ as in Table~\ref{tab:app_batch_regimes}, covering sparse, low-coverage to all-class regimes to test robustness when batch structure is informative but not fully representative. Full implementation details, baseline access patterns, hyperparameters, validation grids, backbone sensitivity, and adaptation-cost measurements appear in Appendix~\ref{app:exp_details} and Tables~\ref{tab:baseline_access_audit_full}--\ref{tab:jpta_compute_cost}.

\textbf{Small-Batch Low-Coverage Regimes:} Table~\ref{tab:jpta_b64_regimes} examines the hardest case: small batches with low effective class coverage, where transductive methods are prone to misinterpreting batch structure. Here, TransCLIP is highly brittle, with average accuracies of $38.9$, $40.4$, and $42.7$ for Very Low, Low, and Medium coverage—well below zero-shot CLIP. This supports StatA’s claim that aggressive transduction can amplify class-support errors in sparse batches.

\begin{figure}[t]
    \centering
    \includegraphics[width=1\linewidth]{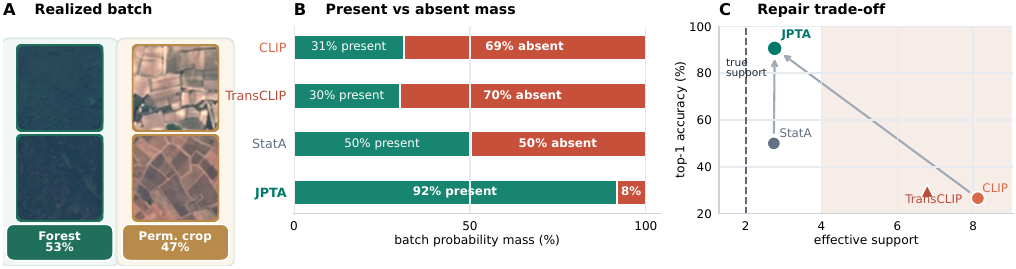}
    \caption{
\textbf{Sparse EuroSAT case study: JPTA suppresses absent-class mass.}
The batch contains only two active classes, \emph{Forest} and \emph{Permanent crop}, while inference is performed over the full label set.
\textbf{A:} Realized batch composition.
\textbf{B:} Batch-averaged posterior mass: CLIP and TransCLIP assign high mass to absent classes (\textcolor{red}{red}), StatA partially reduces this drift, and JPTA concentrates mass on present classes (\textcolor{jptagreen}{green}).
\textbf{C:} JPTA achieves compact predicted support with higher accuracy.
}
\label{fig:eurosat_case_study}
\vspace{-1.8em}
\end{figure}

JPTA performs best in these adverse regimes, reaching 82.4, 80.2, and 76.7 average accuracy across the three $B=64$ settings—margins of +12.1, +9.9, and +9.3 over the strongest baseline. The gains are consistent: JPTA surpasses the strongest baseline in every column and effective-class regime. The largest margins occur on datasets with ambiguous zero-shot predictions and local batch structure, such as DTD, EuroSAT, SUN397, and ImageNet. These results support the central claim that \emph{low-coverage batches require controlled semantic anchoring rather than free batch transduction}; the sparse EuroSAT case in Figure~\ref{fig:eurosat_case_study} further shows that JPTA improves by suppressing absent-class mass and concentrating predictions on the realized batch support.
\begin{table*}[t]
\centering
\caption{
\textbf{Small-batch low effective-class evaluation ($B=64$).}
Very Low, Low, and Medium correspond to $1$--$4$, $2$--$10$, and $5$--$25$ active classes per transductive task, respectively.
Each result is top-1 accuracy (\%) averaged over 1,000 sampled tasks.
CLIP and MTA are regime-independent shared baselines.
For JPTA, the smaller value below each score reports the signed margin $\Delta_{\text{best}}$ relative to the strongest non-JPTA baseline in the same column, including the shared baselines.
}
\label{tab:jpta_b64_regimes}
\vspace{-.25em}
\scriptsize
\setlength{\tabcolsep}{3.3pt}
\renewcommand{\arraystretch}{1.12}
\resizebox{\textwidth}{!}{%
\begin{tabular}{@{}llcccccccccccc@{}}
\toprule
$K_{\mathrm{eff}}$ & Method
& \cellcolor{avgbg}\textbf{Avg}
& ImageNet
& SUN397
& Aircraft
& EuroSAT
& Cars
& Food101
& Pets
& Flowers
& Caltech
& DTD
& UCF101 \\
\midrule

\multirow{2}{*}{}
& CLIP & \fsAvg{65.2} & 66.6 & 62.5 & 24.7 & 48.3 & 65.6 & 85.9 & 89.1 & 70.7 & 93.2 & 43.5 & 67.5 \\
& MTA  & \fsAvg{66.6} & 69.3 & 64.8 & 27.4 & 46.9 & 68.0 & 87.2 & 89.4 & 71.7 & 94.0 & 44.4 & 69.0 \\

\specialrule{0.08em}{0.35em}{0.25em}

\multirow{5}{*}{\makecell[c]{Very Low\\(1--4)}}
& Dirichlet & \fsAvg{68.5} & \fsSecond{79.2} & \fsSecond{75.7} & 28.2 & 47.2 & 68.2 & 88.1 & 87.5 & 71.2 & 88.8 & \fsSecond{50.3} & 69.0 \\
& ZLaP      & \fsAvg{27.5} & 14.5 & 13.0 & 8.4  & 36.6 & 23.7 & 31.9 & 57.0 & 22.4 & 52.4 & 13.0 & 29.2 \\
& TransCLIP & \fsAvg{38.9} & 21.6 & 21.1 & 11.6 & 45.1 & 34.7 & 59.2 & 72.4 & 36.4 & 62.3 & 26.1 & 37.7 \\
& StatA     & \fsAvgSecond{70.4} & 72.9 & 66.0 & \fsSecond{29.3} & \fsSecond{56.8} & \fsSecond{76.2} & \fsSecond{90.3} & \fsSecond{95.5} & \fsSecond{77.6} & 93.0 & 46.1 & \fsSecond{70.2} \\
\rowcolor{oursbg}
& \textbf{JPTA (Ours)}
& \fsAvgJPos{82.4}{12.1}
& \fsJPos{91.5}{12.3}
& \fsJPos{90.0}{14.3}
& \fsJPos{40.1}{10.8}
& \fsJPos{70.6}{13.8}
& \fsJPos{84.2}{8.0}
& \fsJPos{99.5}{9.2}
& \fsJPos{99.5}{4.0}
& \fsJPos{82.9}{5.3}
& \fsJPos{97.7}{3.7}
& \fsJPos{69.3}{19.0}
& \fsJPos{81.6}{11.4} \\

\specialrule{0.08em}{0.35em}{0.25em}

\multirow{5}{*}{\makecell[c]{Low\\(2--10)}}
& Dirichlet & \fsAvgSecond{70.3} & \fsSecond{80.1} & \fsSecond{78.0} & \fsSecond{28.1} & 43.5 & 71.5 & \fsSecond{92.3} & 92.7 & 74.7 & 93.0 & \fsSecond{48.9} & \fsSecond{70.9} \\
& ZLaP      & \fsAvg{35.2} & 19.1 & 19.0 & 12.0 & 46.4 & 27.9 & 43.5 & 66.6 & 31.3 & 60.8 & 22.4 & 38.7 \\
& TransCLIP & \fsAvg{40.4} & 20.3 & 22.4 & 14.3 & \fsSecond{53.9} & 30.8 & 55.6 & 69.4 & 40.9 & 64.6 & 31.6 & 40.9 \\
& StatA     & \fsAvg{69.3} & 72.8 & 66.9 & 27.7 & 51.3 & \fsSecond{73.5} & 89.5 & \fsSecond{93.7} & \fsSecond{76.6} & 93.6 & 46.9 & 69.6 \\
\rowcolor{oursbg}
& \textbf{JPTA (Ours)}
& \fsAvgJPos{80.2}{9.9}
& \fsJPos{90.9}{10.8}
& \fsJPos{85.5}{7.5}
& \fsJPos{36.2}{8.1}
& \fsJPos{66.6}{12.7}
& \fsJPos{82.5}{9.0}
& \fsJPos{98.8}{6.5}
& \fsJPos{98.8}{5.1}
& \fsJPos{80.6}{4.0}
& \fsJPos{97.2}{3.2}
& \fsJPos{64.4}{15.5}
& \fsJPos{80.6}{9.7} \\

\specialrule{0.08em}{0.35em}{0.25em}

\multirow{5}{*}{\makecell[c]{Medium\\(5--25)}}
& Dirichlet & \fsAvgSecond{67.5} & \fsSecond{77.7} & \fsSecond{72.9} & 26.1 & 38.6 & \fsSecond{71.6} & \fsSecond{90.8} & 88.4 & 71.5 & 93.7 & 42.9 & 67.8 \\
& ZLaP      & \fsAvg{44.6} & 29.0 & 27.9 & 16.5 & 49.0 & 36.0 & 59.1 & 76.4 & 42.9 & 72.0 & 32.0 & 50.3 \\
& TransCLIP & \fsAvg{42.7} & 15.5 & 22.8 & 17.0 & \fsSecond{58.2} & 32.9 & 56.3 & 72.6 & 45.0 & 65.6 & 37.5 & 46.5 \\
& StatA     & \fsAvg{67.4} & 70.7 & 65.3 & 26.0 & 45.0 & 71.1 & 88.2 & \fsSecond{90.8} & \fsSecond{73.7} & 93.9 & \fsSecond{47.5} & \fsSecond{69.1} \\
\rowcolor{oursbg}
& \textbf{JPTA (Ours)}
& \fsAvgJPos{76.7}{9.3}
& \fsJPos{87.6}{9.9}
& \fsJPos{82.0}{9.1}
& \fsJPos{32.9}{5.5}
& \fsJPos{58.9}{0.7}
& \fsJPos{80.4}{8.8}
& \fsJPos{96.2}{5.4}
& \fsJPos{97.0}{6.2}
& \fsJPos{78.6}{4.9}
& \fsJPos{96.7}{2.7}
& \fsJPos{57.4}{9.9}
& \fsJPos{76.3}{7.2} \\

\bottomrule
\end{tabular}%
}
\vspace{-2.0em}
\end{table*}
%

\textbf{Large-Batch and Higher-Coverage Regimes:} Table~\ref{tab:jpta_b1000_regimes} tests JPTA as batch size and class coverage grow. As expected, method gaps shrink because batch statistics improve and class-support inflation declines. Still, JPTA attains the best average performance in all three regimes—74.5 (Medium), 73.1 (High), and 71.5 (Very High)—with margins of +4.8, +3.3, and +2.5 over the strongest baseline.

Per-dataset results show nuance. JPTA is not always best: StatA slightly outperforms it on some datasets (e.g., Aircraft, Pets, UCF101, Cars), depending on the regime. These gaps are small and mostly occur where fixed anchoring is already strong. In contrast, JPTA delivers its largest and most consistent gains on ImageNet and EuroSAT, with further gains on SUN397, Flowers, Caltech, and DTD. Food101 shows a strong gain in the Medium regime but narrows as class coverage increases, where batch-aware anchor transport improves class separation without the severe drift seen in TransCLIP. Thus, JPTA is not only effective in the hardest sparse setting; it remains beneficial as batches become more informative.

\begin{table*}[t]
\centering
\caption{
\textbf{Large-batch effective-class evaluation ($B=1000$).}
Medium, High, and Very High correspond to $5$--$25$, $25$--$50$, and $50$--$100$ active classes per transductive task, respectively.
Each result is top-1 accuracy (\%) averaged over 1,000 sampled tasks.
For JPTA, the smaller value below each score reports the signed margin $\Delta_{\text{best}}$ relative to the strongest non-JPTA baseline in the same column.
}
\label{tab:jpta_b1000_regimes}
\vspace{-.25em}
\scriptsize
\setlength{\tabcolsep}{3.3pt}
\renewcommand{\arraystretch}{1.12}
\resizebox{\textwidth}{!}{%
\begin{tabular}{@{}llcccccccccccc@{}}
\toprule
$K_{\mathrm{eff}}$ & Method
& \cellcolor{avgbg}\textbf{Avg}
& ImageNet
& SUN397
& Aircraft
& EuroSAT
& Cars
& Food101
& Pets
& Flowers
& Caltech
& DTD
& UCF101 \\

\specialrule{0.08em}{0.35em}{0.25em}

\multirow{5}{*}{\makecell[c]{Medium\\(5--25)}}
& Dirichlet & \fsAvg{64.4} & 60.9 & \fsSecond{75.4} & \fsSecond{26.7} & 38.8 & \fsSecond{74.1} & 76.2 & 91.0 & 71.6 & 92.4 & 36.2 & 65.4 \\
& ZLaP      & \fsAvg{41.5} & 16.6 & 20.1 & 16.4 & 49.0 & 32.2 & 55.5 & 76.4 & 40.6 & 67.7 & 34.2 & 48.1 \\
& TransCLIP & \fsAvg{56.5} & 39.9 & 42.7 & 22.0 & \fsSecond{63.1} & 49.9 & 80.6 & 87.9 & 58.7 & 79.1 & 42.9 & 55.0 \\
& StatA     & \fsAvgSecond{69.7} & \fsSecond{70.8} & 64.5 & \fsBest{28.4} & 60.4 & 74.0 & \fsSecond{87.5} & \fsBest{93.1} & \fsSecond{77.5} & \fsSecond{92.8} & \fsSecond{47.1} & \fsBest{70.2} \\
\rowcolor{oursbg}
& \textbf{JPTA (Ours)}
& \fsAvgJPos{74.5}{4.8}
& \fsJPos{88.7}{17.9}
& \fsJPos{79.6}{4.2}
& \fsJNeg{26.0}{2.4}
& \fsJPos{68.4}{5.3}
& \fsJPos{77.1}{3.0}
& \fsJPos{94.4}{6.9}
& \fsJNegSecond{92.6}{0.5}
& \fsJPos{79.8}{2.3}
& \fsJPos{94.0}{1.2}
& \fsJPos{50.1}{3.0}
& \fsJNegSecond{68.8}{1.4} \\

\specialrule{0.08em}{0.35em}{0.25em}

\multirow{5}{*}{\makecell[c]{High\\(25--50)}}
& Dirichlet & \fsAvg{45.3} & 17.3 & 37.3 & 21.0 & 37.9 & 65.4 & 46.3 & 81.3 & 46.3 & 80.5 & 21.1 & 43.6 \\
& ZLaP      & \fsAvg{52.2} & 23.8 & 32.2 & 22.2 & 49.3 & 45.4 & 74.9 & 86.5 & 56.2 & 79.7 & 43.6 & 60.8 \\
& TransCLIP & \fsAvg{62.0} & 43.9 & 49.6 & 24.8 & \fsSecond{64.0} & 57.3 & 83.0 & \fsSecond{91.4} & 69.1 & 85.5 & 47.5 & 65.4 \\
& StatA     & \fsAvgSecond{69.7} & \fsSecond{71.9} & \fsSecond{66.4} & \fsSecond{25.9} & 60.7 & \fsSecond{73.6} & \fsSecond{88.0} & \fsSecond{91.4} & \fsSecond{76.7} & \fsSecond{93.2} & \fsSecond{47.9} & \fsBest{71.5} \\
\rowcolor{oursbg}
& \textbf{JPTA (Ours)}
& \fsAvgJPos{73.1}{3.3}
& \fsJPos{84.6}{12.7}
& \fsJPos{71.6}{5.2}
& \fsJPos{26.6}{0.7}
& \fsJPos{69.1}{5.1}
& \fsJPos{73.9}{0.3}
& \fsJPos{90.6}{2.6}
& \fsJPos{92.6}{1.2}
& \fsJPos{79.0}{2.3}
& \fsJPos{95.2}{2.0}
& \fsJPos{50.2}{2.3}
& \fsJNegSecond{70.5}{1.0} \\

\specialrule{0.08em}{0.35em}{0.25em}

\multirow{5}{*}{\makecell[c]{Very High\\(50--100)}}
& Dirichlet & \fsAvg{33.6} & 10.8 & 15.7 & 17.5 & 37.8 & 51.2 & 29.1 & 79.3 & 24.3 & 59.1 & 19.0 & 26.1 \\
& ZLaP      & \fsAvg{58.4} & 32.7 & 44.0 & 25.4 & 49.3 & 55.2 & 83.3 & 87.3 & 64.8 & 87.9 & 45.2 & 67.8 \\
& TransCLIP & \fsAvg{64.4} & 44.5 & 53.0 & \fsSecond{25.6} & \fsSecond{64.1} & 60.9 & 85.2 & \fsSecond{91.9} & \fsSecond{74.3} & 90.5 & \fsSecond{48.1} & \fsSecond{70.7} \\
& StatA     & \fsAvgSecond{69.0} & \fsSecond{71.8} & \fsSecond{67.1} & 23.9 & 60.7 & \fsBest{70.2} & \fsSecond{87.1} & 91.1 & \fsSecond{74.3} & \fsSecond{93.7} & 48.0 & \fsSecond{70.7} \\
\rowcolor{oursbg}
& \textbf{JPTA (Ours)}
& \fsAvgJPos{71.5}{2.5}
& \fsJPos{79.1}{7.3}
& \fsJPos{69.5}{2.4}
& \fsJPos{26.9}{1.3}
& \fsJPos{69.1}{5.0}
& \fsJNegSecond{69.3}{0.9}
& \fsJPos{87.3}{0.2}
& \fsJPos{92.5}{0.6}
& \fsJPos{76.4}{2.1}
& \fsJPos{94.5}{0.8}
& \fsJPos{50.3}{2.2}
& \fsJPos{71.6}{0.9} \\

\bottomrule
\end{tabular}%
}
\end{table*}
%
%

\textbf{Favorable All-Class Evaluation:} Table~\ref{tab:jpta_all_classes} reports results in the full-dataset all-class setting, where the batch label distribution closely matches the full label space and favors standard transductive refinement. This setting evaluates whether JPTA’s robustness in sparse batches harms performance when transduction is reliable. JPTA achieves the best average accuracy of $73.3$, exceeding TransCLIP by $3.0$ points and StatA by $3.4$ points. These gains over StatA are important because both methods use semantic anchoring, but only JPTA adapts the anchor itself; the improvement therefore isolates the value of transported, prior-aware anchoring rather than merely stronger regularization.

JPTA outperforms the strongest baseline on all datasets in the all-class setting, with especially large gains on DTD, UCF101, Flowers, Cars, EuroSAT, and SUN397. This indicates that batch-aware anchoring does not just make the method conservative; it also enables stronger adaptation when batches provide reliable class evidence. Thus, the results address a key robustness concern: JPTA improves sparse-regime stability without reducing accuracy under favorable transductive conditions. 

\begin{table*}[t]
\centering
\caption{
\textbf{All-class transductive evaluation.}
All classes appear in the evaluation batch, i.e., the favorable full-label-space setting.
Results are reported as top-1 accuracy (\%).
CLIP and MTA serve as shared baselines.
For JPTA, the smaller value under each score is the signed margin $\Delta_{\text{best}}$ relative to the strongest non-JPTA baseline in the same column, including the shared baselines.
}
\label{tab:jpta_all_classes}
\scriptsize
\setlength{\tabcolsep}{3.3pt}
\renewcommand{\arraystretch}{1.12}
\resizebox{\textwidth}{!}{%
\begin{tabular}{@{}llcccccccccccc@{}}
\toprule
$K_{\mathrm{eff}}$ & Method
& \cellcolor{avgbg}\textbf{Avg}
& ImageNet
& SUN397
& Aircraft
& EuroSAT
& Cars
& Food101
& Pets
& Flowers
& Caltech
& DTD
& UCF101 \\
\midrule

\multirow{2}{*}{}
& CLIP & \fsAvg{65.2} & 66.6 & 62.5 & 24.7 & 48.3 & 65.6 & 85.9 & 89.1 & 70.7 & 93.2 & 43.5 & 67.5 \\
& MTA  & \fsAvg{66.6} & 69.3 & 64.8 & 27.4 & 46.9 & 68.0 & 87.2 & 89.4 & 71.7 & 94.0 & 44.4 & 69.0 \\

\specialrule{0.08em}{0.35em}{0.25em}

\multirow{5}{*}{All}
& Dirichlet & \fsAvg{29.5} & 66.6 & 0.7 & 15.5 & 39.9 & 21.2 & 8.5 & 76.3 & 14.3 & 44.0 & 19.4 & 18.4 \\
& ZLaP      & \fsAvg{66.4} & 68.8 & 67.2 & 26.9 & 49.1 & 67.1 & 86.9 & 87.8 & 68.7 & 90.8 & 45.5 & 71.4 \\
& TransCLIP & \fsAvgSecond{70.3} & \fsSecond{70.4} & \fsSecond{68.9} & \fsSecond{26.9} & 66.1 & \fsSecond{69.5} & 87.1 & \fsSecond{92.5} & \fsSecond{76.5} & 92.7 & \fsSecond{48.6} & \fsSecond{74.1} \\
& StatA     & \fsAvg{69.9} & 69.9 & 68.7 & 24.7 & \fsSecond{67.3} & 68.0 & 87.1 & 92.4 & 75.2 & \fsSecond{94.2} & 48.4 & 73.5 \\
\rowcolor{oursbg}
& \textbf{JPTA (Ours)}
& \fsAvgJPos{73.3}{3.0}
& \fsJPos{72.3}{1.9}
& \fsJPos{72.1}{3.2}
& \fsJPos{28.2}{0.8}
& \fsJPos{70.6}{3.3}
& \fsJPos{72.9}{3.4}
& \fsJPos{87.5}{0.3}
& \fsJPos{93.4}{0.9}
& \fsJPos{80.0}{3.5}
& \fsJPos{95.1}{0.9}
& \fsJPos{54.2}{5.6}
& \fsJPos{79.6}{5.5} \\

\bottomrule
\end{tabular}%
}
\end{table*}
%
%

\textbf{Online Test-Time Adaptation:} Table~\ref{tab:jpta_online_tta_b128} evaluates online test-time adaptation with correlated streams and class-sequential batches, a more realistic setting than independent batches because consecutive batches can have biased, non-stationary class priors. TENT, TDA, and DMN provide modest gains over zero-shot CLIP, while StatA is more stable due to its fixed semantic anchor.

JPTA outperforms all baselines in every online scenario, exceeding the strongest competitor by $+4.9$, $+5.4$, $+5.7$, and $+6.0$ points for Low, Medium, High, and Separate streams. Gains are especially large on EuroSAT, DTD, UCF101, SUN397, Food101, and Cars, indicating JPTA remains effective under changing batch priors. The Separate scenario is most revealing: classes arrive sequentially, so methods must avoid over-committing to transient evidence. JPTA performs best here, validating regime-adaptive trust and prior-aware refinement for non-i.i.d. test streams.
%
%
\begin{table*}[t]
\centering
\caption{
\textbf{Online test-time adaptation under correlated streams ($B=128$).}
Low, Medium, and High denote stream correlation settings with $\gamma=0.1$, $\gamma=0.01$, and $\gamma=0.001$, respectively; Separate denotes class-sequential streams.
CLIP and MTA are shared, scenario-independent baselines.
Each value is top-1 accuracy (\%) averaged over 100 sampled tasks.
For JPTA, the smaller value below each score is the signed margin $\Delta_{\text{best}}$ relative to the strongest non-JPTA baseline in that column. 
}
\label{tab:jpta_online_tta_b128}
\vspace{-0.25em}
\scriptsize
\setlength{\tabcolsep}{3.3pt}
\renewcommand{\arraystretch}{1.12}
\resizebox{\textwidth}{!}{%
\begin{tabular}{@{}llcccccccccccc@{}}
\toprule
Scenario & Method
& \cellcolor{avgbg}\textbf{Avg}
& ImageNet
& SUN397
& Aircraft
& EuroSAT
& Cars
& Food101
& Pets
& Flowers
& Caltech
& DTD
& UCF101 \\
\midrule

\multirow{2}{*}{Shared}
& CLIP & \fsAvg{65.2} & 66.6 & 62.5 & 24.7 & 48.3 & 65.6 & 85.9 & 89.1 & 70.7 & 93.2 & 43.5 & 67.5 \\
& MTA  & \fsAvg{66.6} & 69.3 & 64.8 & 27.4 & 46.9 & 68.0 & 87.2 & 89.4 & 71.7 & 94.0 & 44.4 & 69.0 \\

\specialrule{0.08em}{0.35em}{0.25em}

\multirow{5}{*}{\makecell[c]{Low\\$(\gamma=0.1)$}}
& TENT  & \fsAvg{65.8} & 66.6 & 64.5 & 24.6 & 51.8 & 65.7 & 85.9 & 89.3 & 70.6 & 93.4 & 44.0 & 67.8 \\
& TDA   & \fsAvg{67.7} & 68.3 & 66.0 & 25.4 & 60.6 & 66.9 & 86.1 & 89.6 & 72.5 & 93.4 & 45.5 & 71.0 \\
& DMN   & \fsAvg{67.2} & 68.0 & 64.8 & 24.9 & 59.8 & 67.0 & 84.2 & 89.9 & 73.3 & 92.2 & 44.8 & 70.3 \\
& StatA & \fsAvg{67.0} & 66.2 & 63.6 & 24.3 & 52.3 & 67.4 & 88.0 & 92.5 & 72.7 & 94.2 & 46.8 & 68.8 \\
\rowcolor{oursbg}
& \textbf{JPTA (Ours)}
& \fsAvgJPos{72.6}{4.9}
& \fsJPos{73.6}{4.3}
& \fsJPos{69.2}{3.2}
& \fsJPos{28.7}{1.3}
& \fsJPos{69.4}{8.8}
& \fsJPos{73.1}{5.1}
& \fsJPos{92.9}{4.9}
& \fsJPos{94.3}{1.8}
& \fsJPos{76.9}{3.6}
& \fsJPos{96.4}{2.2}
& \fsJPos{49.7}{2.9}
& \fsJPos{74.7}{3.7} \\

\specialrule{0.08em}{0.35em}{0.25em}

\multirow{5}{*}{\makecell[c]{Medium\\$(\gamma=0.01)$}}
& TENT  & \fsAvg{65.5} & 66.7 & 64.3 & 24.6 & 47.9 & 65.6 & 85.9 & 89.4 & 70.6 & 93.3 & 44.0 & 67.8 \\
& TDA   & \fsAvg{67.1} & 68.2 & 65.6 & 25.2 & 56.5 & 66.5 & 85.8 & 89.3 & 72.6 & 93.5 & 45.2 & 70.1 \\
& DMN   & \fsAvg{66.5} & 68.0 & 64.8 & 24.9 & 56.2 & 66.8 & 81.9 & 89.0 & 73.0 & 92.1 & 44.9 & 69.6 \\
& StatA & \fsAvg{68.9} & 69.6 & 65.9 & 27.3 & 52.3 & 73.2 & 89.1 & 94.6 & 75.6 & 94.3 & 46.8 & 69.7 \\
\rowcolor{oursbg}
& \textbf{JPTA (Ours)}
& \fsAvgJPos{74.3}{5.4}
& \fsJPos{75.3}{5.7}
& \fsJPos{71.2}{5.3}
& \fsJPos{29.8}{2.4}
& \fsJPos{66.1}{9.6}
& \fsJPos{78.4}{5.2}
& \fsJPos{94.1}{5.0}
& \fsJPos{96.5}{1.9}
& \fsJPos{78.4}{2.8}
& \fsJPos{97.9}{3.6}
& \fsJPos{52.5}{5.7}
& \fsJPos{76.9}{6.8} \\

\specialrule{0.08em}{0.35em}{0.25em}

\multirow{5}{*}{\makecell[c]{High\\$(\gamma=0.001)$}}
& TENT  & \fsAvg{65.3} & 66.8 & 64.3 & 24.8 & 45.6 & 65.6 & 86.1 & 89.4 & 70.5 & 93.4 & 44.0 & 67.9 \\
& TDA   & \fsAvg{66.8} & 67.9 & 65.1 & 25.1 & 55.3 & 66.3 & 85.5 & 89.0 & 72.5 & 93.6 & 45.1 & 69.7 \\
& DMN   & \fsAvg{66.3} & 67.9 & 64.8 & 24.9 & 56.3 & 66.8 & 79.9 & 88.9 & 72.9 & 92.1 & 44.8 & 69.4 \\
& StatA & \fsAvg{69.5} & 71.9 & 66.0 & 27.9 & 51.8 & 74.7 & 89.3 & 94.8 & 76.4 & 94.4 & 47.0 & 69.8 \\
\rowcolor{oursbg}
& \textbf{JPTA (Ours)}
& \fsAvgJPos{75.2}{5.7}
& \fsJPos{76.5}{4.6}
& \fsJPos{72.4}{6.4}
& \fsJPos{31.0}{3.1}
& \fsJPos{66.7}{10.4}
& \fsJPos{79.2}{4.5}
& \fsJPos{95.3}{6.0}
& \fsJPos{96.9}{2.1}
& \fsJPos{80.2}{3.8}
& \fsJPos{98.4}{4.0}
& \fsJPos{53.8}{6.8}
& \fsJPos{77.1}{7.3} \\

\specialrule{0.08em}{0.35em}{0.25em}

\multirow{5}{*}{Separate}
& TENT  & \fsAvg{64.5} & 66.7 & 64.2 & 24.7 & 37.0 & 65.6 & 86.1 & 89.3 & 70.8 & 93.4 & 43.9 & 67.9 \\
& TDA   & \fsAvg{66.6} & 67.4 & 64.6 & 24.9 & 55.3 & 65.9 & 85.2 & 88.9 & 72.3 & 93.6 & 45.0 & 69.6 \\
& DMN   & \fsAvg{65.8} & 67.7 & 64.7 & 24.9 & 55.1 & 66.7 & 78.5 & 88.0 & 72.8 & 91.9 & 44.8 & 69.0 \\
& StatA & \fsAvg{69.1} & 71.7 & 64.9 & 28.9 & 48.2 & 75.2 & 88.9 & 95.2 & 77.6 & 94.3 & 45.8 & 69.0 \\
\rowcolor{oursbg}
& \textbf{JPTA (Ours)}
& \fsAvgJPos{75.1}{6.0}
& \fsJPos{76.3}{4.6}
& \fsJPos{71.8}{6.9}
& \fsJPos{30.8}{1.9}
& \fsJPos{66.9}{11.6}
& \fsJPos{80.4}{5.2}
& \fsJPos{92.1}{3.2}
& \fsJPos{97.2}{2.0}
& \fsJPos{81.4}{3.8}
& \fsJPos{97.7}{3.4}
& \fsJPos{53.4}{7.6}
& \fsJPos{77.8}{8.2} \\
\bottomrule
\end{tabular}%
}
\vspace{-1.5em}
\end{table*}

%
%

\textbf{Dissecting Batch-Aware Anchoring (Ablation Studies):} 
Table~\ref{tab:jpta_ablation_summary} ablates JPTA’s main components. Replacing the batch-aware anchor with the original fixed one causes the largest drop: $10.7$ on average and over $13$ in the very-low coverage regime, indicating that fixed semantic anchoring is too rigid when the batch provides reliable corrective evidence. Transported-only anchoring also underperforms Full JPTA, indicating that transport must be blended with the original semantic reference rather than replacing it. Removing prior-aware refinement or disabling regime-adaptive control causes large losses in sparse regimes, confirming that class-marginal regularization and evidence-dependent trust are needed to avoid class-support inflation. Fixed pseudocounts yield smaller but consistent drops, showing adaptive anchored statistics mainly improve cross-regime stability.

Additional ablations and diagnostics are provided in Appx.~\ref{app:component_ablations} and 
Appx.~\ref{app:mechanism_audit}, including the full per-dataset component study in Table~\ref{tab:jpta_component_ablations}, the anchor/statistic-update
audit in Table~\ref{tab:jpta_mechanism_audit_full}, the oracle diagnostic study in Table~\ref{tab:jpta_oracle_bounds}, and the transport-geometry audit in Table~\ref{tab:jpta_transport_geometry}. 
Qualitative and diagnostic analyses in Appx.~\ref{app:finegrained_case_study} and Appx.~\ref{app:evidence_conditioned_audit} further show how JPTA corrects fine-grained support dilution and applies different adaptive components under different evidence regimes; see Figs~\ref{fig:finegrained_case_study} and~\ref{fig:prior_and_control}. We discuss scope, failure modes, and extensions to open-set or feature-adaptive settings in Appx.~\ref{app:limitations}.

\begin{table*}[t]
\centering
\caption{
\textbf{Compact ablation summary.}
Results are regime-level means over Aircraft, EuroSAT, and Oxford Flowers.
Entries are top-1 accuracy (\%); parentheses show change relative to Full JPTA.
StatA-style uses a fixed CLIP text anchor, Transported-only removes adaptive blending, w/o prior removes batch-marginal refinement, w/o control disables regime-adaptive trust, and Fixed pseudocount replaces adaptive anchored-statistic updates with a fixed schedule.
}
\vspace{-.25em}
\label{tab:jpta_ablation_summary}
\setlength{\tabcolsep}{2.0pt}
\footnotesize
\begin{tabular}{lcccccc}
\toprule
Regime 
& Full JPTA 
& StatA-style 
& Transported-only
& w/o prior 
& w/o control 
& Fixed pseudo. \\
\midrule
Very low $(1$--$4), B{=}64$
& \textbf{64.56}
& 51.47 {\scriptsize($-13.09$)}
& 59.82 {\scriptsize($-4.74$)}
& 61.86 {\scriptsize($-2.70$)}
& 56.46 {\scriptsize($-8.10$)}
& 61.04 {\scriptsize($-3.52$)} \\

Low $(2$--$10), B{=}64$
& \textbf{61.15}
& 51.51 {\scriptsize($-9.64$)}
& 57.59 {\scriptsize($-3.56$)}
& 49.27 {\scriptsize($-11.88$)}
& 50.41 {\scriptsize($-10.74$)}
& 57.14 {\scriptsize($-4.01$)} \\

High $(25$--$50), B{=}1000$
& \textbf{58.28}
& 49.21 {\scriptsize($-9.07$)}
& 55.34 {\scriptsize($-2.94$)}
& 54.48 {\scriptsize($-3.80$)}
& 56.11 {\scriptsize($-2.17$)}
& 56.63 {\scriptsize($-1.65$)} \\

All classes
& \textbf{59.62}
& 48.72 {\scriptsize($-10.90$)}
& 56.28 {\scriptsize($-3.34$)}
& 53.01 {\scriptsize($-6.61$)}
& 55.02 {\scriptsize($-4.60$)}
& 56.35 {\scriptsize($-3.27$)} \\
\midrule
Overall mean
& \textbf{60.90}
& 50.23 {\scriptsize($-10.67$)}
& 57.26 {\scriptsize($-3.64$)}
& 54.66 {\scriptsize($-6.24$)}
& 54.50 {\scriptsize($-6.40$)}
& 57.79 {\scriptsize($-3.11$)} \\
\bottomrule
\end{tabular}
\vspace{-1.5em}
\end{table*}

\noindent\begin{minipage}{\linewidth}
\textbf{Generalization without target-domain retuning:}
We transfer the original CLIP-selected JPTA configuration unchanged to OpenCLIP ViT-B/16 (DataComp-1B)~\cite{gadre2023datacomp} and SigLIP2 So400m/14-384~\cite{tschannen2025siglip2}. Table~\ref{tab:jpta_modern_summary} reports gains over each backbone's own zero-shot baseline across 11 datasets and three regimes. Stronger pretraining reduces headroom but retains adaptation gains; SigLIP2 improves in 29/33 conditions, with four regressions. With ImageNet-selected hyperparameters frozen on ImageNet-A/R/Sketch/V2, JPTA reaches 79.46\% versus StatA's 63.46\% in Very-Low coverage and 69.33\% versus 62.01\% in High coverage (Appx.~\ref{app:source_target_transfer}).

\begin{center}
\centering
\captionsetup{hypcap=false}
\captionof{table}{\textbf{Transfer to modern VLMs without backbone-specific retuning.} Entries are mean accuracy gains (pp) over the same-backbone zero-shot model; each regime averages 11 datasets, and Overall includes Very Low, Very High, and All-Class.}
\label{tab:jpta_modern_summary}
\small
\setlength{\tabcolsep}{8pt}
\begin{tabular}{lcccc}
\toprule
Backbone & Very Low & All-Class & Overall & Improved \\
\midrule
OpenCLIP DataComp-1B & +10.46 & +4.06 & +6.13 & 33/33 \\
SigLIP2 So400m/14-384 & +5.96 & +2.89 & +3.71 & 29/33 \\
\bottomrule
\end{tabular}
\end{center}
Twenty hyperparameter perturbations incur at most 1.33 pp macro loss (Appx.~\ref{app:random_sensitivity}). Paired uncertainty, initialization robustness, and large-batch scaling are detailed in Appx.~\ref{app:paired_statistics}, \ref{app:initialization_robustness}, and~\ref{app:large_batch_scaling}; these analyses also expose negative-adaptation cases.
\end{minipage}

%% file: sec/5_conclusion.tex
\section{Conclusion}
\label{sec:conclusion}
\vspace{-0.5em}
We study a key failure mode of transductive adaptation in vision–language models: the tension between exploiting batch structure and preserving reliable semantic priors. Existing methods make this trade-off explicit: aggressive transduction can hallucinate class support in sparse or prior-shifted batches, while fixed anchoring can ignore useful batch evidence. JPTA resolves this by making the semantic anchor batch-aware. It estimates the batch prior, transports text prototypes toward trustworthy image structure, and refines assignments through prior-aware, regime-adaptive control. Across standard benchmarks, low effective-class regimes, all-class evaluation, and online streams, JPTA consistently outperforms TransCLIP and StatA, with the largest gains in challenging regimes while remaining competitive when batches are favorable. This supports a broader principle for test-time VLM adaptation: robustness should come not from choosing between frozen semantics and free transduction, but from adapting the semantic reference to calibrated batch evidence.


%% file: sec/6_appendix.tex
\clearpage
\appendix
\section{Shrinkage-Controlled Transport and Semantic Drift}
\label{app:shrinkage_drift}
\paragraph{Proof of Proposition 1:}
Let \(s(\eta_k)=\tilde t_k(\eta_k)^\top u_k\). Since
\(\angle(\tilde t_k(\eta_k),u_k)\) decreases when \(s(\eta_k)\) increases, it is
sufficient to show that \(s'(0)>0\). Writing
\(A=\cos\alpha_k\), \(B=\cos\beta_k\), and \(C=\cos\theta_k\), direct
differentiation gives
\[
s'(0)=B-AC .
\]
Because \(\beta_k<\alpha_k<\pi/2\), we have \(B>A>0\), and since \(C\leq 1\),
\(B-AC>0\). Hence, for sufficiently small \(\eta_k>0\),
\(\tilde t_k(\eta_k)\) is closer to \(u_k\) than \(t_k\). The drift expression
follows by computing the dot product
\[
\tilde t_k(\eta)^\top t_k
=
\frac{1-\eta_k+\eta_k\cos\theta_k}
{\|(1-\eta_k)t_k+\eta_k\hat t_k\|},
\]
and using the two-dimensional geometry of the span of \(t_k\) and
\(\hat t_k\).

In the implementation, \(\eta_k\) is calculated from the aggregate soft transport support and the orthogonal alignment score, rather than being learned from labels. Therefore, Proposition~1 explains why the transported anchor can improve alignment when the batch estimate is reliable, while shrinkage limits semantic drift when the transport estimate is weak.

\section{Experimental Protocol and Implementation Details}
\label{app:exp_details}

This appendix provides the full experimental protocol, baseline coverage, implementation details, and hyperparameter selection strategy used in Sec.~\ref{sec:experiments}. Unless stated otherwise, all methods are evaluated under the same frozen CLIP feature space, text prompts, batch construction, and random seeds.

\subsection{Datasets and Feature Extraction}
\label{app:datasets}

We follow the transductive VLM evaluation protocol of TransCLIP~\cite{zanella2024boosting} and StatA~\cite{zanella2025realistic}. Experiments are run on ImageNet~\cite{deng2009imagenet} and ten transfer benchmarks: Caltech101~\cite{fei2004learning}, OxfordPets~\cite{parkhi2012cats}, StanfordCars~\cite{krause20133d}, Flowers102~\cite{nilsback2008automatedflowers}, Food101~\cite{bossard2014food}, FGVC-Aircraft~\cite{maji2013fineaircraft}, SUN397~\cite{xiao2010sun}, DTD~\cite{cimpoi2014describing}, EuroSAT~\cite{helber2019eurosat}, and UCF101~\cite{soomro2012ucf101}, spanning object recognition, fine-grained classification, textures, scenes, actions, food, and remote sensing.

All main experiments use the frozen CLIP ViT-B/16 image encoder and its corresponding frozen text encoder; all methods share the same cached image features and normalized text prototypes. Image features are extracted once and cached before transductive inference. Text prototypes are obtained by encoding class-specific prompts with the frozen text encoder and $\ell_2$-normalizing the embeddings. Unless stated otherwise, we use the TransCLIP/StatA prompt templates and class-name processing. No method sees test labels during adaptation.

For fair comparison, all transductive methods use the same normalized image features, normalized text prototypes, and unlabeled batches. Performance differences therefore stem only from the transductive solver, not feature extraction, prompt engineering, or batching.

\subsection{Batch Construction and Effective-Class Regimes}
\label{app:batch_construction}

We evaluate methods under three regimes: independent batch adaptation, low effective-class adaptation, and online adaptation.

\paragraph{Independent batch adaptation.}
Each test batch is adapted independently. At the start of each batch, the transductive solver is reset and given only that batch’s unlabeled features. Predictions are made after batch-level refinement. This measures how well each method exploits the current batch without using information from other batches.

\paragraph{Low effective-class adaptation.}
To test robustness under realistic class-prior shift, we vary the effective number of active classes per batch. For a dataset with $K$ classes, we sample $K_{\mathrm{eff}}$ active classes and form a batch of size $B$ using only those classes. The global label space is fixed, but each batch may cover only a small subset. This reveals the failure mode highlighted in StatA~\cite{zanella2025realistic}: transductive methods may perform well when all classes appear, but degrade when batches are sparse or prior-shifted.

\paragraph{Online adaptation.}
We also evaluate realistic online streams when applicable. Test batches arrive sequentially and may have correlated class composition. Each method processes the stream without labels. Stateless methods operate on each batch independently, while online-capable methods update internal statistics according to their protocol. All methods see the same batch order.

\begin{table}[t]
\centering
\caption{
Effective-class regimes used for realistic transductive evaluation. 
For datasets with fewer than the upper-bound number of classes, $K_{\mathrm{eff}}$ is clipped by the dataset class count $K$.
}
\label{tab:app_batch_regimes}
\small
\begin{tabular}{lcc}
\toprule
Regime & Batch size $B$ & Effective classes $K_{\mathrm{eff}}$ \\
\midrule
Very Low & 64 & 1--4 \\
Low & 64 & 2--10 \\
Medium & 64 & 5--25 \\
Medium & 1000 & 5--25 \\
High & 1000 & 25--50 \\
Very High & 1000 & 50--100 \\
All Classes & full test set / all-class batch & $K$ \\
\bottomrule
\end{tabular}
\end{table}

All reported results are averaged over repeated random batch samplings and random seeds. The same sampled batches are reused across methods. We report top-1 accuracy as the primary metric. 

\subsection{Baselines}
\label{app:baselines}

We compare JPTA against four groups of baselines.

\begin{table*}[t]
\centering
\caption{
\textbf{Baseline coverage and access-pattern audit for VLM test-time and transductive adaptation.}
A method is considered a direct protocol match only if it uses an unlabeled test batch, requires no
labeled support examples, keeps CLIP encoders/prompts frozen, and performs adaptation through a
frozen-feature batch-level transductive solver.
}
\label{tab:baseline_access_audit_full}
\vspace{0.3em}
\scriptsize
\setlength{\tabcolsep}{3.5pt}
\renewcommand{\arraystretch}{1.10}

\begin{tabular}{@{}p{0.14\textwidth}p{0.25\textwidth}ccccc@{}}
\toprule
\textbf{Method}
& \textbf{Adaptation family}
& \makecell{\textbf{Batch}\\\textbf{eval.}}
& \makecell{\textbf{Online}\\\textbf{eval.}}
& \makecell{\textbf{No support}\\\textbf{labels}}
& \makecell{\textbf{Frozen}\\\textbf{CLIP}}
& \makecell{\textbf{Direct}\\\textbf{match}} \\
\midrule

\rowcolor{oursbg}
\multicolumn{7}{@{}l}{\textit{Shared reference baselines}} \\
CLIP \cite{radford2021learningclip}
& Zero-shot reference
& \cmark & \cmark & \cmark & \cmark & \xmark \\
MTA \cite{zanella2024testMTA}
& Static reference baseline
& \cmark & \cmark & \cmark & \cmark & \xmark \\

\midrule
\rowcolor{oursbg}
\multicolumn{7}{@{}l}{\textit{Frozen-feature transductive baselines used in the main batch evaluation}} \\
Dirichlet \cite{martin2024transductive}
& Prior-based transductive refinement
& \cmark & \xmark & \cmark & \cmark & \cmark \\
ZLaP \cite{kalantidis2024labelZLap}
& Graph label propagation
& \cmark & \xmark & \cmark & \cmark & \cmark \\
TransCLIP \cite{zanella2024boosting}
& Free batch transduction
& \cmark & \xmark & \cmark & \cmark & \cmark \\
StatA \cite{zanella2025realistic}
& Fixed-anchor transduction
& \cmark & \cmark & \cmark & \cmark & \cmark \\

\midrule
\rowcolor{oursbg}
\multicolumn{7}{@{}l}{\textit{Online or memory-based test-time adaptation baselines}} \\
TENT \cite{wang2020tent}
& Entropy-minimization TTA
& \xmark & \cmark & \cmark & \xmark & \xmark \\
TDA \cite{karmanov2024efficient}
& Dynamic cache adaptation
& \xmark & \cmark & \cmark & \cmark & \xmark \\
DMN \cite{zhang2024dual}
& Dual-memory adaptation
& \xmark & \cmark & \cmark & \cmark & \xmark \\
FreeTTA \cite{dai2025free}
& Online distribution adaptation
& \xmark & \xmark & \cmark & \cmark & \xmark \\
BCA \cite{zhou2025bayesian}
& Bayesian class/prior adaptation
& \xmark & \xmark & \cmark & \xmark & \xmark \\
DPE \cite{DPE}
& Dual prototype evolving
& \xmark & \xmark & \cmark & \cmark & \xmark \\
BoostAdapter \cite{zhang2024boostadapter}
& Regional bootstrapping and memory adaptation
& \xmark & \xmark & \cmark & \cmark & \xmark \\
COSMIC \cite{huang2025cosmic}
& Cross-modal semantic cache adaptation
& \xmark & \xmark & \cmark & \xmark & \xmark \\
BATCLIP \cite{maharana2025batclip}
& Bimodal online test-time adaptation
& \xmark & \xmark & \cmark & \xmark & \xmark \\
CLIPTTA \cite{lafon2025cliptta}
& Contrastive feature/model adaptation
& \xmark & \xmark & \cmark & \xmark & \xmark \\
WATT \cite{osowiechi2024watt}
& Weight-averaged test-time adaptation
& \xmark & \xmark & \cmark & \xmark & \xmark \\

\midrule
\rowcolor{oursbg}
\multicolumn{7}{@{}l}{\textit{Prompt-, attribute-, or language-augmented test-time adaptation methods}} \\
TPT \cite{shu2022test}
& Test-time prompt tuning
& \xmark & \xmark & \cmark & \xmark & \xmark \\
DiffTPT \cite{feng2023diverse}
& Diffusion-augmented prompt tuning
& \xmark & \xmark & \cmark & \xmark & \xmark \\
PromptAlign \cite{abdul2023align}
& Distribution-aligned test-time prompting
& \xmark & \xmark & \cmark & \xmark & \xmark \\
C-TPT \cite{yoon2024ctpt}
& Calibrated test-time prompt tuning
& \xmark & \xmark & \cmark & \xmark & \xmark \\
O-TPT \cite{sharifdeen2025otpt}
& Orthogonality-constrained prompt tuning
& \xmark & \xmark & \cmark & \xmark & \xmark \\
A-TPT \cite{ahamed2026atpt}
& Angular-diversity prompt calibration
& \xmark & \xmark & \cmark & \xmark & \xmark \\
R-TPT \cite{sheng2025rtpt}
& Robust test-time prompt tuning
& \xmark & \xmark & \cmark & \xmark & \xmark \\
SCAP \cite{zhang2025scap}
& Clique-based attribute prompting
& \xmark & \xmark & \cmark & \xmark & \xmark \\
GTA-CLIP \cite{saha2025generate}
& Generate--transduct--adapt with attributes
& \xmark & \xmark & \cmark & \xmark & \xmark \\
LIMO \cite{baklouti2026limo}
& Language-aware transductive few-shot CLIP
& \xmark & \xmark & \xmark & \xmark & \xmark \\
OTTER \cite{shin2024otter}
& Label-distribution adaptation
& \xmark & \xmark & \cmark & \cmark & \xmark \\

\midrule
\textbf{JPTA}
& \textbf{Batch-conditioned semantic anchoring}
& \cmark & \cmark & \cmark & \cmark & \cmark \\

\bottomrule
\end{tabular}

\vspace{0.35em}
\begin{minipage}{0.96\textwidth}
\footnotesize
\textbf{Notes.}
“Direct match” assumes unlabeled test-batch access, no labeled support data, frozen CLIP encoders/prompts, and adaptation only via a frozen-feature, batch-level transductive solver. CLIP and MTA serve as non-adaptive baselines. TENT, TDA, and DMN are used in the online stream setting as test-time adaptation baselines. Other approaches—prompt-, feature-, weight-, memory-, or few-shot adaptation—are related but do not match the exact access pattern considered by JPTA.
\end{minipage}

\vspace{-0.6em}
\end{table*}

\paragraph{Zero-shot reference.}
\textbf{Zero-shot CLIP} classifies each image via cosine similarity between frozen image embeddings and frozen text prototypes. It is the non-adaptive reference, indicating whether test-time adaptation improves or harms the pretrained VLM.

\paragraph{Transductive VLM baselines.}
The main baselines are \textbf{TransCLIP}~\cite{zanella2024boosting} and \textbf{StatA}~\cite{zanella2025realistic}. TransCLIP performs strong batch-level transductive refinement by exploiting unlabeled batch structure and the text prior. StatA improves robustness by anchoring batch adaptation to the original text prototypes. They illustrate the core trade-off in our work: free transduction adapts strongly but can drift on sparse batches, while fixed anchoring is stable but may under-adapt when the batch provides strong evidence.

\paragraph{Classical transductive and feature-space baselines.}
When compatible with our protocol, we include representative transductive baselines from prior work, such as Dirichlet-based refinement, ZLaP, LaplacianShot~\cite{ziko2020laplacian}, TIM~\cite{boudiaf2020information}, and PT-MAP-style prototype correction~\cite{hu2021leveraging}. These methods mainly operate in visual feature space and do not explicitly model the language-side semantic anchor as VLM-specific transductive methods do.

\paragraph{Adjacent test-time VLM adaptation baselines.}
We also evaluate broader test-time VLM adaptation methods whose access patterns match our setting. These include prompt-tuning methods such as TPT~\cite{shu2022test}, C-TPT~\cite{yoon2024ctpt}, O-TPT~\cite{sharifdeen2025otpt}, and variants, as well as unlabeled adaptation methods such as CLIP Reward~\cite{zhao2024clipreward}, UMFC~\cite{liang2024umfc}, WATT~\cite{osowiechi2024watt}, CLIPTTA~\cite{lafon2025cliptta}, and ROSITA~\cite{sreenivas2025rosita}. These methods contextualize JPTA within test-time adaptation, but the main paper centers on TransCLIP and StatA, which share JPTA’s batch-transductive access pattern and frozen CLIP feature space and thus provide the most direct comparisons.

\paragraph{Baseline access pattern.}
Table~\ref{tab:baseline_access_audit_full} summarizes the access patterns of recent VLM test-time and transductive adaptation methods. Our main batch evaluation targets frozen-feature transductive baselines that follow the same unlabeled, no-support-label protocol as JPTA, including Dirichlet, ZLaP, TransCLIP, and StatA. The online evaluation adds stream-based test-time adaptation baselines (TENT, TDA, DMN) to assess robustness under non-i.i.d. sequential batches. Several recent methods are related but do not exactly match our protocol: prompt-based methods update textual prompts at inference; memory/cache methods accumulate information across test streams; Bayesian or label-distribution methods adjust priors or class embeddings; and attribute/LLM-based methods add extra language-side adaptation. We therefore cite and audit these methods for completeness, but restrict primary comparisons to methods with the same frozen-feature, unlabeled-batch transductive access pattern.

\paragraph{Equal-budget paired baseline comparison.}
For the additional Low and Very-High comparisons analyzed in Appx.~\ref{app:paired_statistics}, TransCLIP, StatA, and JPTA use the original TransCLIP/StatA data loaders and share validation splits, sampled images, class subsets, seeds, batch sizes, effective-class ranges, and task manifests. Each method is tuned on the same held-out tasks with the same selection metric and the same number of configurations per regime. Selected settings are frozen before test evaluation; no test labels or per-task hyperparameter selection are used. Test comparisons share frozen CLIP features and prompts, and graph-based methods share the same $k$-NN construction. TransCLIP's sparse-regime degradation therefore persists after equal-budget low-coverage tuning, rather than reflecting only an inherited All-Class configuration. This paired baseline study is distinct from the ImageNet-to-shifted-domain transfer protocol in Appx.~\ref{app:source_target_transfer}.

\subsection{Implementation Details}
\label{app:implementation}

JPTA is a gradient-free transductive solver on frozen CLIP features. It does not fine-tune the image encoder, text encoder, prompts, adapters, or any VLM parameters at test time. For each batch, JPTA initializes from the zero-shot CLIP posterior, estimates batch-level class marginals, builds soft image-side class prototypes, transports text prototypes toward the batch geometry, and iteratively refines assignments and anchored class statistics.

\paragraph{Feature normalization.}
All image and text embeddings are $\ell_2$-normalized before inference, so cosine similarity reduces to an inner product. The CLIP logit scale is fixed and shared across zero-shot CLIP, TransCLIP, StatA, and JPTA.

\paragraph{Graph construction.}
For each batch, we build a $k$-nearest-neighbor graph in normalized image-feature space, used only for transductive inference and reconstructed per batch unless the online protocol states otherwise. The same graph construction is used for all graph-based baselines.

\paragraph{Anchor transport.}
JPTA forms soft image-side class prototypes from the current assignment matrix and transports text prototypes toward them via a weighted alignment step. The transported anchor is blended with the original text anchor using a trust coefficient set by batch evidence: when the batch is ambiguous, sparse, or shows low neighborhood agreement, JPTA stays close to original CLIP semantics; when the batch is coherent, the transported anchor has a stronger influence.

\paragraph{Prior-aware refinement.}
The estimated batch class prior regularizes assignment updates. It limits probability mass on many inactive classes in low-coverage batches, while staying flexible when the batch prior is uncertain. The same estimate also drives the adaptive trust mechanism for anchor transport and anchored statistics.

\paragraph{Numerical stability.}
All probability vectors are clipped by a small constant $\epsilon$ before logarithms or normalization. Covariances use diagonal statistics with variance flooring. Classes with very small effective mass are updated conservatively and remain near their semantic anchor.

\paragraph{Algorithm overview.}
Algorithm~\ref{alg:jpta_inference} summarizes the order of the existing updates, distinguishing quantities computed from the initial batch from those updated during refinement.
\begin{algorithm}[H]
\caption{JPTA: Batch-Conditioned Semantic Anchoring}
\label{alg:jpta_inference}
\small
\begin{algorithmic}[1]
\Require Frozen normalized image features $X=\{x_i\}_{i=1}^N$, text prototypes $T=\{t_k\}_{k=1}^K$, pretrained logit scale $\tau$, graph size $k$, fixed solver hyperparameters, iteration budget $T_{\max}$
\Ensure Predicted labels $\{\hat y_i\}_{i=1}^N$
\State Compute the zero-shot posterior $A^{(0)}$; set $Z\gets A^{(0)}$ \Comment{Eq.~\ref{eq:zeroshot_anchor}}
\State Construct the batch $k$-NN graph $W$ and its Laplacian $L_G$
\State Compute initial batch evidence and initialize the smoothed prior $\pi$ \Comment{Eqs.~\ref{eq:margin}--\ref{eq:soft_coverage}}
\State Initialize anchored class statistics
\For{$t=0,\ldots,T_{\max}-1$}
    \State Compute effective supports $n_k\gets\sum_i z_{ik}$ and soft image prototypes $\bar x_k$
    \State Set confidence weights $w_k$ and evidence-dependent transport trust
    \State $M\gets\sum_k w_k\bar x_k t_k^\top$; compute $M=U S V^\top$
    \State $R^\star\gets UV^\top$; $\hat t_k\gets R^\star t_k$ \Comment{Eq.~\ref{eq:procrustes}}
    \State $\tilde t_k\gets\mathrm{norm}((1-\eta_k)t_k+\eta_k\hat t_k)$ \Comment{Eq.~\ref{eq:procrustes}}
    \State Compute $A^{(\mathrm{tr})}$ and the blended anchor $A^{(\mathrm{anc})}$ \Comment{Eq.~\ref{eq:transport_anchor}}
    \State Hold the current likelihood, anchor, prior, and graph message fixed
    \State Refine assignments $Z$ using the evidence-selected marginal branch \Comment{Eq.~\ref{eq:jpta_update}}
    \State Compute empirical moments $\bar\mu_k,\bar\Sigma_k$ \Comment{Eq.~\ref{eq:empirical_moments}}
    \State Update anchored class statistics $\mu_k,\Sigma_k$ and refresh $\pi$ \Comment{Eqs.~\ref{eq:anchored_updates},~\ref{eq:hard_coverage}}
    \If{the solver's early-stop condition holds}
        \State \textbf{break}
    \EndIf
\EndFor
\State \Return $\hat y_i\gets\arg\max_k z_{ik}$ for each image $i$
\end{algorithmic}
\end{algorithm}
The original posterior, graph, and initial pseudo-label evidence are batch-initial quantities; assignments, soft prototypes, the prior, transport, and anchored moments belong to the alternating refinement. Independent batches reset the solver. Label-informed oracle variants in Appx.~\ref{app:ablations} are diagnostic interventions and are not used by this inference procedure.

\subsection{Hyperparameter Selection}
\label{app:hyperparams}

All hyperparameters are chosen without using test labels. When a validation split is available, we tune them once on that split and fix them for all test batches. When following a prior benchmark, we keep its TransCLIP/StatA settings for shared components (feature extraction, graph construction, batching). We do not perform any per-dataset test-time tuning.

\begin{table}[t]
\centering
\caption{
JPTA hyperparameters. Shared protocol parameters follow TransCLIP/StatA where applicable. JPTA-specific parameters are selected once on validation data and then fixed for test evaluation.
}
\label{tab:app_hyperparams}
\small
\begin{tabular}{lll}
\toprule
Symbol / parameter & Role & Selection protocol \\
\midrule
$B$ & batch size & fixed by regime, e.g., $64$ or $1000$ \\
$K_{\mathrm{eff}}$ & active classes per batch & fixed by regime in Table~\ref{tab:app_batch_regimes} \\
$k$ & nearest neighbors for graph & shared with TransCLIP/StatA protocol \\
$T_{\max}$ & refinement iterations & fixed on validation split \\
$\tau$ & CLIP logit scale & frozen CLIP value \\
$\lambda_a$ & anchor weight & validation-selected \\
$\lambda_g$ & graph consistency weight & validation-selected or inherited from protocol \\
$\lambda_\pi$ & batch-prior regularization weight & validation-selected \\
\bottomrule
\end{tabular}
\end{table}


\begin{table}[t]
\centering
\caption{
\textbf{Fixed JPTA hyperparameters.}
We report the method-level hyperparameters used in the evaluation protocol.
The configuration is selected on validation tasks and then held fixed for test-set evaluation.
Controller thresholds are fixed to default values in all experiments and are not tuned per dataset.
}
\label{tab:jpta_selected_hyperparameters}
\scriptsize
\setlength{\tabcolsep}{5pt}
\begin{tabular}{lcccccc}
\toprule
Regime 
& $k$ 
& $T_{\max}$ 
& $\lambda_a$ 
& $\lambda_g$ 
& $\lambda_\pi$ 
& $\alpha_\pi$ \\
\midrule
Small-batch low coverage & $3$ & $10$ & $1$ & $1$ & $1$ & $10$ \\
Large-batch high coverage & $3$ & $10$ & $1$ & $1$ & $1$ & $10$ \\
All-class & $3$ & $10$ & $1$ & $1$ & $1$ & $10$ \\
Online streams & $3$ & $10$ & $1$ & $1$ & $1$ & $10$ \\
\bottomrule
\end{tabular}
\end{table}

\begin{table}[t]
\centering
\caption{
\textbf{Validation grid for JPTA method-level hyperparameters.}
The final configuration in Table~\ref{tab:jpta_selected_hyperparameters} is selected from this compact validation grid.
}
\label{tab:jpta_validation_grid}
\scriptsize
\setlength{\tabcolsep}{5pt}
\begin{tabular}{lc}
\toprule
Hyperparameter & Search space \\
\midrule
$k$ & $\{3,5,10\}$ \\
$T_{\max}$ & $\{5,10,20\}$ \\
$\lambda_a$ & $\{0.25,0.5,1,2\}$ \\
$\lambda_g$ & $\{0.01,0.05,0.1,0.5,1\}$ \\
$\lambda_\pi$ & $\{0.01,0.05,0.1,0.5,1\}$ \\
$\alpha_\pi$ & $\{1,5,10,20,50\}$ \\
\bottomrule
\end{tabular}
\end{table}

For transparency, we report the hyperparameter search space for JPTA-specific components. Unless an ablation states otherwise, the same selected hyperparameters are used for all datasets. This avoids hidden per-dataset tuning and ensures improvements stem from the proposed transductive mechanism rather than benchmark-specific calibration.

\subsection{Random Hyperparameter Perturbations}
\label{app:random_sensitivity}
To test robustness away from the selected configuration, we evaluate 20 predeclared Latin-hypercube perturbations of the active regularization weights over a $[0.5\times,2\times]$ interval. The same perturbation settings are applied to ten datasets in Very-Low, High, and All-Class regimes, with no validation-based selection among the 20 settings. All other settings, including controller thresholds, prior strength, and iteration budget, remain frozen. This gives 600 perturbed evaluations and 30 frozen-reference evaluations. The pretrained logit scale is fixed for each backbone and shared with its zero-shot baseline; it is not tuned by JPTA.

\begin{center}
\begin{minipage}{\linewidth}
\centering
\captionsetup{hypcap=false}
\captionof{table}{\textbf{Hyperparameter perturbation sensitivity.} Impact is the macro accuracy difference (pp) from the corresponding frozen JPTA reference. SD, 10th percentile, and worst impact summarize the 20 settings, rather than uncertainty over sampled test tasks.}
\label{tab:jpta_random_sensitivity}
\small
\setlength{\tabcolsep}{7pt}
\begin{tabular}{lccc}
\toprule
Regime & Mean impact $\pm$ SD & 10th percentile & Worst impact \\
\midrule
Very Low & $-0.53\pm0.35$ & $-0.98$ & $-1.24$ \\
High & $-0.41\pm0.25$ & $-0.82$ & $-0.85$ \\
All-Class & $-0.76\pm0.37$ & $-1.26$ & $-1.33$ \\
\bottomrule
\end{tabular}
\end{minipage}
\end{center}
The average loss is below one percentage point in every regime, and the worst macro loss among these settings is 1.33 pp. This is evidence of local robustness around the frozen configuration, rather than a guarantee for arbitrary hyperparameters. Individual datasets, particularly EuroSAT, exhibit greater variation than these macro summaries. Source-to-target transfer in Appx.~\ref{app:source_target_transfer} separately tests a configuration selected without target-domain validation.

\subsection{Backbone sensitivity.}
\label{app:Backbone sensitivity.}
Table~\ref{tab:clip_encoder_ablation_b64} tests whether JPTA’s gains depend on a specific CLIP encoder or hold across backbone capacities. Under the same $B=64$ transductive protocol, JPTA outperforms the strongest baseline for both RN50 and ViT-L. Improvements are most pronounced in the sparse Very Low regime, where limited class coverage weakens fixed semantic anchors and batch-aware transported anchors yield a stronger adaptation signal. As expected, ViT-L achieves higher absolute accuracy than RN50, but the relative pattern is consistent: JPTA improves average performance across effective-class regimes, showing that the adaptation mechanism is not tied to a single backbone architecture.

\begin{table*}[htbp]
\centering
\caption{Comparison of various CLIP encoders for the batch test-time adaptation setting. Each reported performance is averaged over 1,000 tasks.
}
\label{tab:clip_encoder_ablation_b64}
\scriptsize
\setlength{\tabcolsep}{3.1pt}
\renewcommand{\arraystretch}{1.12}
\resizebox{\textwidth}{!}{%
\begin{tabular}{@{}llcccccccccccc@{}}
\toprule

\multicolumn{14}{c}{\textbf{(a) RN50}} \\
\midrule
$K_{\mathrm{eff}}$ & Method
& \cellcolor{avgbg}\textbf{Avg}
& ImageNet
& SUN397
& Aircraft
& EuroSAT
& Cars
& Food101
& Pets
& Flowers
& Caltech
& DTD
& UCF101 \\
\midrule

\multirow{1}{*}{}
& CLIP
& \fsAvg{58.7} & 58.2 & 58.9 & 17.0 & 36.2 & 55.8 & 77.4 & 85.7 & 66.1 & 85.7 & 42.8 & 61.8 \\

\specialrule{0.06em}{0.30em}{0.20em}

\multirow{2}{*}{\makecell[c]{Very Low\\(1--4)}}
& StatA
& \fsAvg{65.8} & 68.2 & 63.7 & 21.1 & 43.3 & 71.3 & 87.1 & 93.1 & 74.1 & 90.1 & 45.3 & 66.2 \\
\rowcolor{oursbg}
& \textbf{JPTA (Ours)}
& \fsJPos{73.3}{7.5}
& \fsJPos{79.1}{10.9}
& \fsJPos{76.3}{12.6}
& \fsJPos{25.5}{4.4}
& \fsJPos{45.0}{1.7}
& \fsJPos{80.6}{9.3}
& \fsJPos{96.8}{9.7}
& \fsJPos{95.5}{2.4}
& \fsJPos{75.2}{1.1}
& \fsJPos{95.2}{5.1}
& \fsJPos{54.6}{9.3}
& \fsJPos{82.3}{16.1} \\

\specialrule{0.06em}{0.30em}{0.20em}

\multirow{2}{*}{\makecell[c]{Low\\(2--10)}}
& StatA
& \fsAvg{63.5} & 65.4 & 63.1 & 16.5 & 51.7 & 65.4 & 81.0 & 84.4 & 70.0 & 88.3 & 47.2 & 66.0 \\
\rowcolor{oursbg}
& \textbf{JPTA (Ours)}
& \fsJPos{65.8}{2.3}
& \fsJPos{68.9}{3.5}
& \fsJPos{66.3}{3.2}
& \fsJPos{17.5}{1.0}
& \fsJPos{52.9}{1.2}
& \fsJPos{67.6}{2.2}
& \fsJPos{85.3}{4.3}
& \fsJPos{88.9}{4.5}
& \fsJPos{72.2}{2.2}
& \fsJPos{89.3}{1.0}
& \fsJPos{48.0}{0.8}
& \fsJPos{67.3}{1.3} \\

\specialrule{0.06em}{0.30em}{0.20em}

\multirow{2}{*}{\makecell[c]{Medium\\(5--25)}}
& StatA
& \fsAvg{62.4} & 60.4 & 64.3 & 16.0 & 50.5 & 58.2 & 77.9 & 87.7 & 67.7 & 87.3 & 48.5 & 67.5 \\
\rowcolor{oursbg}
& \textbf{JPTA (Ours)}
& \fsJPos{66.6}{4.2}
& \fsJPos{61.8}{1.4}
& \fsJPos{65.6}{1.3}
& \fsJPos{17.5}{1.5}
& \fsJPos{70.8}{20.3}
& \fsJPos{60.6}{2.4}
& \fsJPos{79.6}{1.7}
& \fsJPos{89.7}{2.0}
& \fsJPos{75.3}{7.6}
& \fsJPos{90.4}{3.1}
& \fsJPos{51.7}{3.2}
& \fsJPos{69.5}{2.0} \\

\specialrule{0.10em}{0.70em}{0.30em}

\multicolumn{14}{c}{\textbf{(b) ViT-L/14}} \\
\midrule
$K_{\mathrm{eff}}$ & Method
& \cellcolor{avgbg}\textbf{Avg}
& ImageNet
& SUN397
& Aircraft
& EuroSAT
& Cars
& Food101
& Pets
& Flowers
& Caltech
& DTD
& UCF101 \\
\midrule

\multirow{1}{*}{}
& CLIP
& \fsAvg{72.6} & 73.5 & 67.7 & 32.5 & 60.3 & 76.9 & 90.9 & 93.5 & 79.5 & 95.2 & 53.5 & 74.9 \\

\specialrule{0.06em}{0.30em}{0.20em}

\multirow{2}{*}{\makecell[c]{Very Low\\(1--4)}}
& StatA
& \fsAvg{77.3} & 78.9 & 71.3 & 40.4 & 71.4 & 84.4 & 94.2 & 97.1 & \textbf{82.9} & 97.0 & 55.3 & 77.1 \\
\rowcolor{oursbg}
& \textbf{JPTA (Ours)}
& \fsJPos{83.1}{5.8}
& \fsJPos{88.5}{9.6}
& \fsJPos{84.5}{13.2}
& \fsJPos{42.8}{2.4}
& \fsJPos{72.9}{1.5}
& \fsJPos{86.7}{2.3}
& \fsJPos{99.3}{5.1}
& \fsJPos{98.8}{1.7}
& \fsJNeg{81.1}{1.8}
& \fsJPos{99.5}{2.5}
& \fsJPos{72.4}{17.1}
& \fsJPos{87.4}{10.3} \\

\specialrule{0.06em}{0.30em}{0.20em}

\multirow{2}{*}{\makecell[c]{Low\\(2--10)}}
& StatA
& \fsAvg{76.3} & 77.2 & 70.9 & 36.8 & 71.2 & 82.0 & 92.3 & 94.3 & 81.9 & 95.3 & 58.7 & 78.8 \\
\rowcolor{oursbg}
& \textbf{JPTA (Ours)}
& \fsJPos{79.8}{3.5}
& \fsJPos{84.2}{7.0}
& \fsJPos{80.1}{9.2}
& \fsJPos{37.9}{1.1}
& \fsJPos{81.1}{9.9}
& \fsJPos{85.3}{3.3}
& \fsJPos{94.3}{2.0}
& \fsJPos{95.1}{0.8}
& \fsJPos{82.8}{0.9}
& \fsJPos{97.5}{2.2}
& \fsJPos{59.6}{0.9}
& \fsJPos{79.8}{1.0} \\

\specialrule{0.06em}{0.30em}{0.20em}

\multirow{2}{*}{\makecell[c]{Medium\\(5--25)}}
& StatA
& \fsAvg{76.7} & 76.9 & \textbf{72.8} & 35.4 & 76.7 & 78.0 & 91.8 & 94.6 & 81.8 & 95.0 & 59.7 & 81.3 \\
\rowcolor{oursbg}
& \textbf{JPTA (Ours)}
& \fsJPos{79.7}{3.0}
& \fsJPos{77.9}{1.0}
& \fsJNeg{71.0}{1.8}
& \fsJPos{37.3}{1.9}
& \fsJPos{86.9}{10.2}
& \fsJPos{81.4}{3.4}
& \fsJPos{92.2}{0.4}
& \fsJPos{95.9}{1.3}
& \fsJPos{88.2}{6.4}
& \fsJPos{97.5}{2.5}
& \fsJPos{65.2}{5.5}
& \fsJPos{83.2}{1.9} \\

\bottomrule
\end{tabular}%
}
\vspace{-0.2em}
\end{table*}

\subsection{Transfer to OpenCLIP and SigLIP2}
\label{app:modern_backbones}
We extend the RN50 and ViT-L/14 analysis to OpenCLIP ViT-B/16 trained on DataComp-1B~\cite{gadre2023datacomp} and SigLIP2 So400m/14-384~\cite{tschannen2025siglip2}. Each backbone is evaluated on all 11 benchmarks in Very-Low, Very-High, and All-Class regimes, giving 33 dataset--regime conditions per backbone and 66 additional evaluations. JPTA uses the configuration selected with the original CLIP ViT-B/16, without backbone-specific retuning. Every gain compares JPTA with the corresponding backbone's own zero-shot predictions under the same evaluation condition; it is not a comparison against zero-shot OpenAI CLIP.

Table~\ref{tab:jpta_modern_summary} summarizes the macro gains. OpenCLIP DataComp-1B improves in 33/33 conditions by 6.13 pp overall; SigLIP2 improves in 29/33 by 3.71 pp. The Very-Low gains are 10.46 and 5.96 pp, respectively. In All-Class evaluation, each backbone improves on all 11 datasets, by 4.06 and 2.89 pp on average. SigLIP2's mean zero-shot accuracy across the 33 conditions is already 81.95\%, yet useful adaptation headroom remains. The four SigLIP2 regressions occur in sampled regimes, and gains tend to narrow when the initial accuracy is already high. These comparisons establish benefit over the same-backbone zero-shot reference, without asserting superiority over untested modern-backbone adaptation baselines or predicting the behavior of future VLMs.

\begin{center}
\begin{minipage}{\linewidth}
\centering
\captionsetup{hypcap=false}
\captionof{table}{\textbf{Fine-grained Very-Low gains with modern backbones.} Accuracy improvements (pp) are relative to the same-backbone zero-shot model, using the transferred JPTA configuration.}
\label{tab:jpta_modern_finegrained}
\small
\setlength{\tabcolsep}{9pt}
\begin{tabular}{lcc}
\toprule
Dataset & OpenCLIP DataComp-1B & SigLIP2 So400m/14-384 \\
\midrule
FGVC-Aircraft & +9.22 & +8.57 \\
Stanford Cars & +7.67 & +2.71 \\
Oxford Pets & +5.47 & +2.12 \\
\bottomrule
\end{tabular}
\end{minipage}
\end{center}
These fine-grained results complement the Stanford Cars case study in Appx.~\ref{app:finegrained_case_study}: visually similar but absent classes can dilute support even when the backbone is strong. They do not imply recovery when the initial semantic correspondence is uninformative; Appx.~\ref{app:initialization_robustness} documents that boundary.

\subsection{Source-to-Target Transfer under Natural Domain Shift}
\label{app:source_target_transfer}
We select the complete configuration on ImageNet and transfer it unchanged to ImageNet-A~\cite{hendrycks2021naturalImagenet-A}, ImageNet-R~\cite{hendrycks2021manyImagenet-r}, ImageNet-Sketch~\cite{wang2019learningImagenet-Sketch}, and ImageNet-V2~\cite{recht2019imagenetv2}, without using target-domain validation labels. Table~\ref{tab:jpta_natural_shift} reports macro accuracy across the four target datasets. JPTA outperforms CLIP and StatA on every shifted dataset in both Very-Low and High coverage; the paired 95\% confidence intervals for these comparisons exclude zero. It improves over CLIP on 87.1\% of Very-Low tasks and 95.9\% of High-coverage tasks.

\begin{center}
\begin{minipage}{\linewidth}
\centering
\captionsetup{hypcap=false}
\captionof{table}{\textbf{Natural domain shift with frozen ImageNet-selected hyperparameters.} Entries are macro top-1 accuracy (\%) over ImageNet-A/R/Sketch/V2. The best method in each row is bold.}
\label{tab:jpta_natural_shift}
\small
\setlength{\tabcolsep}{12pt}
\begin{tabular}{lccc}
\toprule
Regime & Zero-shot CLIP & StatA & JPTA \\
\midrule
Very Low & 59.51 & 63.46 & \textbf{79.46} \\
High & 59.71 & 62.01 & \textbf{69.33} \\
\bottomrule
\end{tabular}
\end{minipage}
\end{center}
The shifted All-Class result is more modest: JPTA improves over CLIP by 0.67 pp on average but trails TransCLIP by 0.70 pp. Thus, the strongest gains occur under partial class coverage; favorable all-class conditions do not guarantee that JPTA is the best adaptation method on every domain.

\subsection{Paired Statistical Analysis}
\label{app:paired_statistics}
We rerun CLIP, StatA, TransCLIP, and JPTA on identical task manifests and compute paired statistics from task-level or image-level predictions. These are additional paired reruns; their precise gains need not equal differences computed from rounded entries in the original main tables. We distinguish a fixed comparator from a dataset-wise strongest comparator. In Table~\ref{tab:jpta_paired_statistics}, the latter is the best of CLIP, StatA, and TransCLIP by mean accuracy within each dataset and regime; it is not a different comparator selected for each test task.

For regime-level macro inference, each of the 11 datasets contributes one paired mean accuracy difference. The confidence intervals, paired $t$-tests, and Wilcoxon signed-rank tests in Table~\ref{tab:jpta_paired_statistics} use these 11 dataset-level differences. We do not pool 11,000 sampled tasks as independent observations for this macro analysis. Separately, within each sampled dataset we analyze its 1,000 paired task differences and apply Holm correction across the 11 dataset-wise primary comparisons. All-Class within-dataset comparisons use paired per-image predictions and exact McNemar tests, again with Holm correction across datasets.

\begin{center}
\begin{minipage}{\linewidth}
\centering
\captionsetup{hypcap=false}
\captionof{table}{\textbf{Paired macro gains over the dataset-wise strongest comparator.} Gains and 95\% CIs are in pp. Macro tests use 11 dataset-level differences; the final column counts datasets with positive mean gains.}
\label{tab:jpta_paired_statistics}
\small
\setlength{\tabcolsep}{5pt}
\begin{tabular}{lccccc}
\toprule
Regime & Gain & 95\% CI & Paired $t$: $p$ & Wilcoxon: $p$ & Positive \\
\midrule
Low & +9.871 & [5.775, 13.967] & 0.0003 & 0.0010 & 11/11 \\
Very High & +2.306 & [0.448, 4.164] & 0.0199 & 0.0049 & 10/11 \\
All-Class & +2.714 & [1.546, 3.882] & 0.0004 & 0.0010 & 11/11 \\
\bottomrule
\end{tabular}
\end{minipage}
\end{center}
Using a single comparator for the regime, the paired reruns give +2.92 pp over StatA in Very High (95\% CI [0.82, 5.02]) and +2.98 pp over TransCLIP in All-Class ([1.83, 4.12]). These differ from the stricter dataset-wise comparison above, which allows the strongest comparator to vary by dataset.

In Low coverage, all 11 dataset-wise gains remain significant after Holm correction, ranging from +3.50 pp on Oxford Pets to +19.58 pp on SUN397. In Very High, 10/11 datasets improve after correction; Stanford Cars is 0.472 pp below StatA, with paired effect size $d_z=-0.254$. In All-Class, all 11 improvements remain significant after Holm-corrected exact McNemar tests, including Food101 (+0.406 pp) and Caltech101 (+0.771 pp). Statistical significance does not imply uniform benefit on individual tasks; task-level harm is examined in Appx.~\ref{app:initialization_robustness}.

\subsection{Reproducibility}
\label{app:reproducibility}

All experiments use fixed random seeds for batch sampling and solver initialization. For each dataset and regime, all methods evaluate the same batches. Image features and text prototypes are computed once and reused across baselines. For each run, we log the CLIP backbone, prompt template, dataset split, batch size, effective-class range, random seed, and solver hyperparameters.

JPTA adds no trainable parameters to the frozen VLM and needs no test-time gradient updates. Its extra cost comes from batch-level graph construction, soft prototype computation, anchor transport, and iterative assignment refinement. Because all operations occur in the cached CLIP feature space, this overhead is small compared to feature extraction. Runtime and memory results appear in Table~\ref{tab:jpta_compute_cost}.

\paragraph{Adaptation cost.}
JPTA runs after the shared CLIP image and text encoders. To isolate the adaptation cost, we time JPTA on cached ViT-B/16 image features and cached text anchors, using the same sampled-batch protocol as the main experiments. As shown in Table~\ref{tab:jpta_compute_cost}, JPTA takes about 6.6--7.5 ms/image for sparse $B{=}64$ batches and amortizes to about 0.54 ms/image at $B{=}1000$ (about 1.8k adapted images/second). Thus, the transported-anchor and controller computations add only a small post-feature cost and scale well with batch size.

\begin{table}[t]
\centering
\small
\setlength{\tabcolsep}{5pt}
\caption{
\textbf{JPTA adaptation cost across batch regimes.}
Timing uses cached ViT-B/16 image features and cached text anchors, excludes the shared CLIP forward pass, and reports median wall-clock time over sampled batches on a single RTX 4090 Laptop GPU. Peak memory reports the maximum GPU memory allocated during adaptation.
}
\label{tab:jpta_compute_cost}
\begin{tabular}{lccrc}
\toprule
Regime & Batch time (ms) & ms/image & Throughput & Peak mem. \\
\midrule
1--4, $B{=}64$      & 419.9 & 6.561 & 152 img/s  & 115.1 MB \\
2--10, $B{=}64$     & 445.2 & 6.956 & 144 img/s  & 115.1 MB \\
5--25, $B{=}64$     & 482.2 & 7.534 & 133 img/s  & 115.1 MB \\
5--25, $B{=}1000$   & 535.2 & 0.535 & 1868 img/s & 358.0 MB \\
25--50, $B{=}1000$  & 545.8 & 0.546 & 1832 img/s & 412.7 MB \\
50--100, $B{=}1000$ & 543.1 & 0.543 & 1841 img/s & 414.5 MB \\
\bottomrule
\end{tabular}
\vspace{-0.5cm}
\end{table}

\paragraph{Existing assets.}
All experiments use publicly available datasets and pretrained CLIP models. We cite the original papers for ImageNet, Caltech101, OxfordPets, StanfordCars, Flowers102, Food101, FGVC-Aircraft, SUN397, DTD, EuroSAT, UCF101, and CLIP. We do not redistribute these datasets; users must obtain them and the pretrained models from their official sources. 

\subsection{Large-Batch Sinkhorn and Graph-Update Scaling}
\label{app:large_batch_scaling}
We profile cached ImageNet-1K CLIP features while retaining all $K=1000$ prototypes, $k=3$, and $T_{\max}=10$. To isolate the expensive branch, we enable exact Sinkhorn normalization at every assignment update and exercise the full iteration budget. Each run performs 55 Sinkhorn normalizations and 55 sparse graph-propagation updates. Table~\ref{tab:jpta_large_batch_scaling} is therefore a stress test of the fully activated branch, distinct from the evidence-routed timing in Table~\ref{tab:jpta_compute_cost}.

\begin{center}
\begin{minipage}{\linewidth}
\centering
\captionsetup{hypcap=false}
\captionof{table}{\textbf{Forced exact-Sinkhorn scaling on ImageNet-1K features.} Times are medians of five measured runs after three warm-ups on an NVIDIA RTX 4090 Laptop GPU. Shared feature extraction is excluded; all 1,000 class prototypes are retained.}
\label{tab:jpta_large_batch_scaling}
\small
\setlength{\tabcolsep}{7pt}
\begin{tabular}{rcccc}
\toprule
Batch size & Sinkhorn (s) & Graph updates (s) & Total JPTA (s) & ms/image \\
\midrule
1,000 & 0.349 & 0.046 & 0.742 & 0.742 \\
2,000 & 0.364 & 0.052 & 0.756 & 0.378 \\
4,000 & 0.415 & 0.073 & 0.848 & 0.212 \\
8,000 & 0.628 & 0.115 & 1.184 & 0.148 \\
16,000 & 2.279 & 0.241 & 3.129 & 0.196 \\
\bottomrule
\end{tabular}
\end{minipage}
\end{center}
At $B=16{,}000$, Sinkhorn and iterative graph propagation together account for 80.6\% of total adaptation time. One Sinkhorn normalization of the $B\times K$ assignment matrix costs $O(T_{\mathrm{SK}}BK)$ time and $O(BK)$ working memory, with $T_{\mathrm{SK}}=25$. Given a sparse $k$-NN graph with $O(Bk)$ edges, one propagation update costs $O(BkK)$. This propagation bound assumes the graph is already constructed and is not a bound on neighbor search or the complete solver. Normal execution may use the softer KL marginal penalty or terminate earlier, so these measurements characterize the forced branch on the stated hardware rather than universal end-to-end latency.

\section{Additional Ablations and Qualitative Analysis}
\label{app:ablations}

This appendix provides full ablation results supporting Sec.~\ref{sec:experiments}. The main paper gives only a compact regime-level summary for space. Here, we present per-dataset component ablations, a mechanism audit of the anchor and statistic-update design, and qualitative case studies showing how JPTA corrects effective-class mismatch in sparse and fine-grained batches. Unless noted, all ablations reuse the hyperparameters of Full JPTA, and batch-regime comparisons are paired over the same sampled transductive tasks.

\subsection{Full Component Ablations}
\label{app:component_ablations}

Table~\ref{tab:jpta_component_ablations} presents per-dataset component ablations on FGVC-Aircraft, EuroSAT, and Oxford Flowers. These datasets stress distinct failure modes: fine-grained ambiguity, domain-specific label structure, and visually confusable classes. We test each ablation in sparse small-batch, higher-coverage large-batch, and all-class regimes.

The results show that the transported semantic anchor and regime-adaptive control are essential in sparse regimes. Removing the transported anchor reduces JPTA to a fixed-anchor variant and sharply degrades performance on all datasets, especially EuroSAT and Oxford Flowers. Disabling adaptive control is similarly harmful in low-coverage batches, indicating that fixed transport and anchor weights fail when batch evidence varies across tasks. In contrast, prior-aware refinement has regime-dependent impact: it is crucial on EuroSAT, where class-support inflation is severe, but less important in some high-coverage fine-grained settings. Overall, the pattern supports JPTA as a selective adaptation framework rather than a uniformly stronger update rule.

\begin{table*}[t]
\centering
\scriptsize
\caption{
\textbf{Full component ablations for JPTA.}
We isolate the contribution of transported semantic anchoring, prior-aware refinement, and regime-adaptive control on FGVC-Aircraft, EuroSAT, and Oxford Flowers across sparse, high-coverage, and all-class regimes.
Each entry reports top-1 accuracy (\%); values in parentheses denote the change relative to Full JPTA under the same dataset and regime.
All ablations reuse the hyperparameters selected for the full model.
Batch-regime comparisons are paired over the same 1,000 sampled tasks, while all-class rows use the full test split.
}
\label{tab:jpta_component_ablations}
\setlength{\tabcolsep}{6pt}
\begin{tabular}{llcccc}
\toprule
Dataset & Regime & \fsAvg{Full JPTA} & Fixed anchor (StatA-style) & w/o prior & w/o control \\
\midrule
FGVC-Aircraft & Very low (1--4), B=64 & \fsAvg{\textbf{40.10}} & 29.36 {\scriptsize(\nd{-10.74})} & 38.45 {\scriptsize(\nd{-1.65})} & 27.40 {\scriptsize(\nd{-12.70})} \\
FGVC-Aircraft & Low (2--10), B=64 & \fsAvg{\textbf{36.24}} & 28.57 {\scriptsize(\nd{-7.67})} & 32.72 {\scriptsize(\nd{-3.52})} & 26.32 {\scriptsize(\nd{-9.92})} \\
FGVC-Aircraft & High (25--50), B=1000 & \fsAvg{\textbf{26.67}} & 21.54 {\scriptsize(\nd{-5.13})} & 27.10 {\scriptsize(\pd{0.43})} & 23.97 {\scriptsize(\nd{-2.70})} \\
FGVC-Aircraft & All classes & \fsAvg{\textbf{28.23}} & 18.23 {\scriptsize(\nd{-10.00})} & 27.06 {\scriptsize(\nd{-1.17})} & 25.68 {\scriptsize(\nd{-2.55})} \\
\midrule
EuroSAT & Very low (1--4), B=64 & \fsAvg{\textbf{70.63}} & 57.47 {\scriptsize(\nd{-13.16})} & 66.87 {\scriptsize(\nd{-3.76})} & 64.40 {\scriptsize(\nd{-6.23})} \\
EuroSAT & Low (2--10), B=64 & \fsAvg{\textbf{66.61}} & 57.74 {\scriptsize(\nd{-8.87})} & 37.87 {\scriptsize(\nd{-28.74})} & 54.35 {\scriptsize(\nd{-12.26})} \\
EuroSAT & High (25--50), B=1000 & \fsAvg{\textbf{69.14}} & 59.10 {\scriptsize(\nd{-10.04})} & 56.91 {\scriptsize(\nd{-12.23})} & 67.65 {\scriptsize(\nd{-1.49})} \\
EuroSAT & All classes & \fsAvg{\textbf{70.64}} & 61.69 {\scriptsize(\nd{-8.95})} & 55.07 {\scriptsize(\nd{-15.57})} & 65.12 {\scriptsize(\nd{-5.52})} \\
\midrule
Oxford Flowers & Very low (1--4), B=64 & \fsAvg{\textbf{82.94}} & 67.58 {\scriptsize(\nd{-15.36})} & 80.27 {\scriptsize(\nd{-2.67})} & 77.59 {\scriptsize(\nd{-5.35})} \\
Oxford Flowers & Low (2--10), B=64 & \fsAvg{\textbf{80.61}} & 68.22 {\scriptsize(\nd{-12.39})} & 77.21 {\scriptsize(\nd{-3.40})} & 70.57 {\scriptsize(\nd{-10.04})} \\
Oxford Flowers & High (25--50), B=1000 & \fsAvg{\textbf{79.04}} & 66.98 {\scriptsize(\nd{-12.06})} & 79.43 {\scriptsize(\pd{0.39})} & 76.72 {\scriptsize(\nd{-2.32})} \\
Oxford Flowers & All classes & \fsAvg{\textbf{79.98}} & 66.23 {\scriptsize(\nd{-13.75})} & 76.90 {\scriptsize(\nd{-3.08})} & 74.26 {\scriptsize(\nd{-5.72})} \\
\bottomrule
\end{tabular} 
\end{table*}
%
%
\subsection{Anchor Blending and Anchored-Statistic Updates}
\label{app:mechanism_audit}

Table~\ref{tab:jpta_mechanism_audit_full} analyzes the mechanisms used to construct and stabilize the batch-aware anchor. We compare Full JPTA with three variants: \emph{Original-only}, which keeps the fixed CLIP text anchor; \emph{Transported-only}, which uses the transported anchor without adaptive blending; and \emph{Fixed pseudocount}, which replaces the adaptive anchored-statistic schedule with a fixed one.

Original-only consistently underperforms Full JPTA, showing that fixed text semantics are too rigid when the batch provides reliable corrections. Transported-only performs substantially better than original-only but still lags Full JPTA, indicating that transport helps but unregulated replacement of the pretrained anchor is less robust than controlled interpolation. Fixed-pseudocount yields smaller but consistent losses, suggesting that adaptive anchored statistics mainly improve stability across regimes. Overall, JPTA’s benefit arises from combining transport, semantic preservation, and adaptive statistical trust, not from transport alone.

\definecolor{JPTADrop}{HTML}{B00020}
\newcommand{\drop}[1]{\textcolor{JPTADrop}{#1}}

\begin{table*}[t]
\centering
\caption{
\textbf{Anchor and statistic-update audit.}
We compare Full JPTA with three mechanism variants: Original-only, which uses the fixed CLIP text anchor; Transported-only, which uses the transported anchor without adaptive blending; and Fixed pseudocount, which removes adaptive anchored-statistic scheduling.
Each entry reports top-1 accuracy (\%); values in parentheses denote the change relative to Full JPTA.
The final rows report regime-level means over FGVC-Aircraft, EuroSAT, and Oxford Flowers.
}
\label{tab:jpta_mechanism_audit_full}
\setlength{\tabcolsep}{3.8pt}
\footnotesize
\begin{tabular}{llcccc}
\toprule
Dataset & Regime & Full JPTA & Original-only (StatA-style) & Transported-only & Fixed pseudocount \\
\midrule
FGVC-Aircraft & Very low (1--4), B=64
& \textbf{40.10} & 29.36 {\scriptsize(\drop{-10.74})}
& 36.42 {\scriptsize(\drop{-3.68})}
& 37.88 {\scriptsize(\drop{-2.22})} \\
FGVC-Aircraft & Low (2--10), B=64
& \textbf{36.24} & 28.57 {\scriptsize(\drop{-7.67})}
& 33.41 {\scriptsize(\drop{-2.83})}
& 34.52 {\scriptsize(\drop{-1.72})} \\
FGVC-Aircraft & High (25--50), B=1000
& \textbf{26.67} & 21.54 {\scriptsize(\drop{-5.13})}
& 25.42 {\scriptsize(\drop{-1.25})}
& 25.86 {\scriptsize(\drop{-0.81})} \\
FGVC-Aircraft & All classes
& \textbf{28.23} & 18.23 {\scriptsize(\drop{-10.00})}
& 25.86 {\scriptsize(\drop{-2.37})}
& 26.91 {\scriptsize(\drop{-1.32})} \\
\midrule

EuroSAT & Very low (1--4), B=64
& \textbf{70.63} & 57.47 {\scriptsize(\drop{-13.16})}
& 66.12 {\scriptsize(\drop{-4.51})}
& 66.74 {\scriptsize(\drop{-3.89})} \\
EuroSAT & Low (2--10), B=64
& \textbf{66.61} & 57.74 {\scriptsize(\drop{-8.87})}
& 62.93 {\scriptsize(\drop{-3.68})}
& 59.82 {\scriptsize(\drop{-6.79})} \\
EuroSAT & High (25--50), B=1000
& \textbf{69.14} & 59.10 {\scriptsize(\drop{-10.04})}
& 65.41 {\scriptsize(\drop{-3.73})}
& 67.12 {\scriptsize(\drop{-2.02})} \\
EuroSAT & All classes
& \textbf{70.64} & 61.69 {\scriptsize(\drop{-8.95})}
& 66.88 {\scriptsize(\drop{-3.76})}
& 65.31 {\scriptsize(\drop{-5.33})} \\
\midrule

Oxford Flowers & Very low (1--4), B=64
& \textbf{82.94} & 67.58 {\scriptsize(\drop{-15.36})}
& 76.93 {\scriptsize(\drop{-6.01})}
& 78.49 {\scriptsize(\drop{-4.45})} \\
Oxford Flowers & Low (2--10), B=64
& \textbf{80.61} & 68.22 {\scriptsize(\drop{-12.39})}
& 76.44 {\scriptsize(\drop{-4.17})}
& 77.09 {\scriptsize(\drop{-3.52})} \\
Oxford Flowers & High (25--50), B=1000
& \textbf{79.04} & 66.98 {\scriptsize(\drop{-12.06})}
& 75.20 {\scriptsize(\drop{-3.84})}
& 76.90 {\scriptsize(\drop{-2.14})} \\
Oxford Flowers & All classes
& \textbf{79.98} & 66.23 {\scriptsize(\drop{-13.75})}
& 76.10 {\scriptsize(\drop{-3.88})}
& 76.84 {\scriptsize(\drop{-3.14})} \\
\midrule

Mean & Very low (1--4), B=64
& \textbf{64.56} & 51.47 {\scriptsize(\drop{-13.09})}
& 59.82 {\scriptsize(\drop{-4.74})}
& 61.04 {\scriptsize(\drop{-3.52})} \\
Mean & Low (2--10), B=64
& \textbf{61.15} & 51.51 {\scriptsize(\drop{-9.64})}
& 57.59 {\scriptsize(\drop{-3.56})}
& 57.14 {\scriptsize(\drop{-4.01})} \\
Mean & High (25--50), B=1000
& \textbf{58.28} & 49.21 {\scriptsize(\drop{-9.07})}
& 55.34 {\scriptsize(\drop{-2.94})}
& 56.63 {\scriptsize(\drop{-1.65})} \\
Mean & All classes
& \textbf{59.62} & 48.72 {\scriptsize(\drop{-10.90})}
& 56.28 {\scriptsize(\drop{-3.34})}
& 56.35 {\scriptsize(\drop{-3.27})} \\
\bottomrule
\end{tabular}
\vspace{2pt}
\begin{minipage}{0.98\linewidth}
\footnotesize
\emph{Note:} Original-only uses the fixed CLIP text anchor. Transported-only uses the transported anchor without adaptive blending. Fixed pseudocount replaces the adaptive anchored-statistic schedule with a fixed one.
\end{minipage}
\vspace{-0.5cm}
\end{table*}
\subsection{Fine-Grained Case Study on Stanford Cars}
\label{app:finegrained_case_study}

\begin{figure}
    \centering
    \includegraphics[width=1\linewidth]{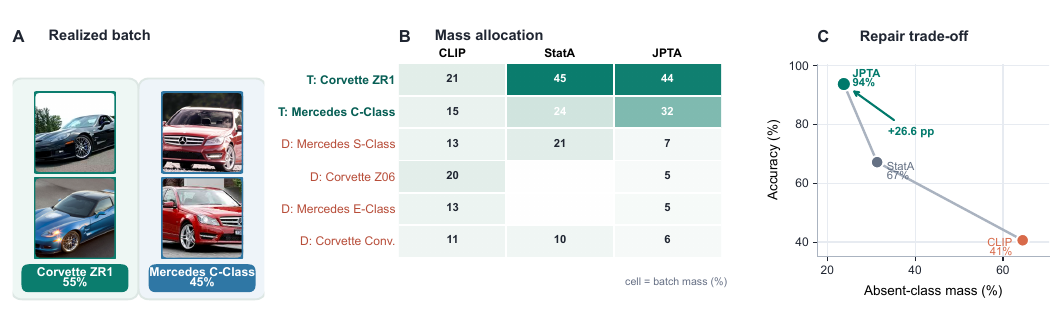}
    \caption{\textbf{Fine-grained support dilution explains why statistic-only adaptation is insufficient.}
\textbf{A:} A low-coverage Stanford Cars batch contains only two realized subtypes: Corvette ZR1 (\textcolor{jptagreen}{green}) and Mercedes C-Class (\textcolor{blue}{blue}).
\textbf{B:} Heatmap entries show batch-averaged class probability mass. True batch labels are marked in \textcolor{jptagreen}{green} and absent distractor labels in \textcolor{orange}{orange}. CLIP spreads probability over visually plausible but absent fine-grained labels; StatA increases mass on one true class but still assigns substantial mass to distractors. JPTA reallocates mass toward the realized support, especially the second true subtype, while suppressing absent alternatives.
\textbf{C:} This yields the desired repair trade-off: JPTA moves to lower absent-class mass and higher accuracy, improving over StatA by $26.6$ percentage points on the same batch.}
\label{fig:finegrained_case_study}
\vspace{-0.5cm}
\end{figure}

Figure~\ref{fig:finegrained_case_study} shows a low-coverage Stanford Cars batch with two fine-grained subtypes. This is challenging because visually similar but absent classes act as plausible distractors. CLIP spreads probability mass over these distractors, and StatA only partly corrects this. JPTA instead reallocates mass to the realized classes—especially the weaker true subtype—while suppressing absent alternatives. This shows that statistic-only adaptation is insufficient in fine-grained regimes: the semantic anchor must move toward the observed batch geometry, but under controlled trust.

\subsection{Evidence-Conditioned Mechanism Audit}
\label{app:evidence_conditioned_audit}

\begin{figure}[htbp]
    \centering
    \includegraphics[width=\linewidth]{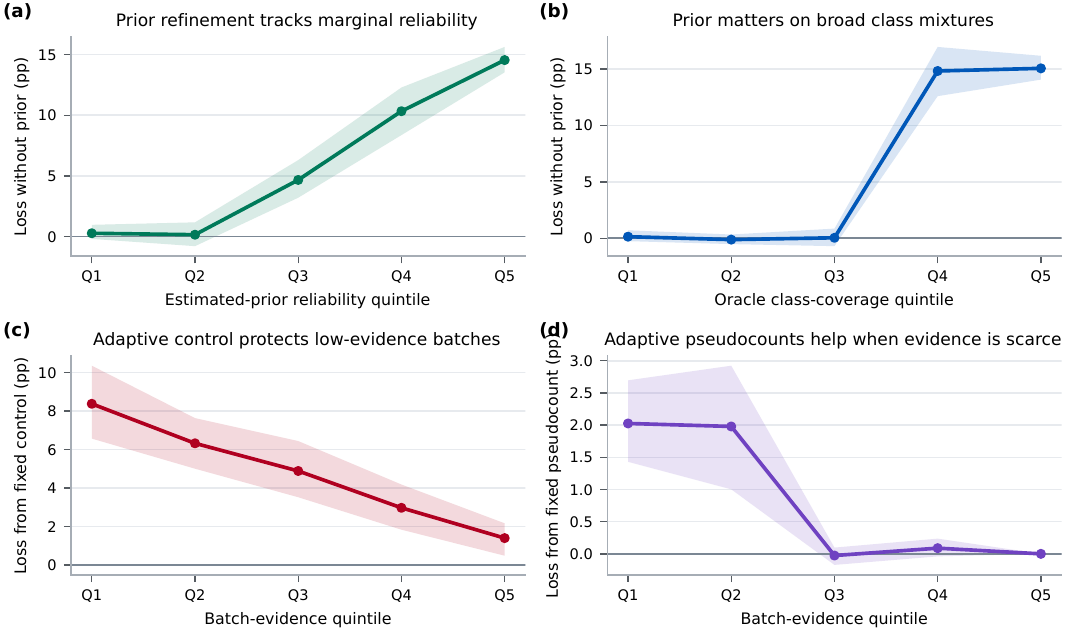}
    \caption{
    \textbf{Evidence-conditioned ablation analysis.}
    We sort paired transductive tasks into quintiles using diagnostics computed by Full JPTA and report the accuracy loss caused by each ablation within each quintile.
    Prior-aware refinement is most valuable when the estimated batch marginal is reliable and class coverage is broader.
    Adaptive control and adaptive pseudocounts matter most in low-evidence batches, where fixed schedules are more prone to drift.
    Bands show bootstrap 95\% confidence intervals over paired task batches.
    }
    \label{fig:prior_and_control}
\end{figure}

Figure~\ref{fig:prior_and_control} examines when each adaptive component matters by grouping paired transductive tasks into diagnostic quintiles and measuring accuracy loss from ablating each component. Prior-aware refinement helps most when the estimated batch marginal is reliable and class coverage is broad, while adaptive control and adaptive pseudocounts matter most for low-evidence batches where fixed schedules are unstable. This shows that JPTA does not merely strengthen transduction, but adapts selectively to batch evidence.

\subsection{Oracle diagnostics.}
Table~\ref{tab:jpta_oracle_bounds} uses label-informed oracles only as diagnostics, with all other hyperparameters fixed. Oracle support and oracle prior yield substantial gains, especially in low-coverage regimes, while oracle trust is essentially identical to Full JPTA. Thus, remaining headroom is not due to a poorly tuned controller: the learned trust schedule already matches its label-informed counterpart. The main bottleneck is instead estimating batch support and class marginals from unlabeled test features—the support-mismatch problem JPTA targets.

\begin{table}[htbp]
\centering
\small
\setlength{\tabcolsep}{4.2pt}
\caption{\textbf{Oracle diagnostics for JPTA.}
Results are regime-level means over FGVC-Aircraft, EuroSAT, and Oxford Flowers.
Accuracies are percentages; values in parentheses denote the change relative to Full JPTA in the same regime.
Oracle variants use label information only for analysis, with all other hyperparameters fixed:
oracle prior gives the true batch class marginal, oracle support gives the set of present classes without per-sample labels,
and oracle trust uses labels only to choose the batch-level trust schedule.
The large oracle-prior and oracle-support gains show that JPTA is mainly limited by estimating batch support and marginals,
while the near-zero oracle-trust gap indicates that the learned controller is already close to its label-informed diagnostic schedule.}
\label{tab:jpta_oracle_bounds}
\begin{tabular}{lccccc}
\toprule
Regime & ZS & Full JPTA & + oracle prior & + oracle support & + oracle trust \\
\midrule
Very low (1--4), $B{=}64$
& 50.21 & \textbf{64.19}
& 72.66 (\textcolor{jptagreen}{+8.47})
& 81.90 (\textcolor{jptagreen}{+17.71})
& 64.15 (\textcolor{gray}{-0.04}) \\
High (25--50), $B{=}1000$
& 49.48 & \textbf{58.24}
& 65.01 (\textcolor{jptagreen}{+6.77})
& 68.44 (\textcolor{jptagreen}{+10.20})
& 58.22 (\textcolor{gray}{-0.02}) \\
All classes
& 49.68 & \textbf{60.13}
& 65.89 (\textcolor{jptagreen}{+5.76})
& 60.13 (\textcolor{jptagray}{0.00})
& 60.11 (\textcolor{gray}{-0.02}) \\
\midrule
Overall mean
& 49.79 & \textbf{60.85}
& 67.85 (\textcolor{jptagreen}{+7.00})
& 73.49 (\textcolor{jptagreen}{+12.64})
& 60.83 (\textcolor{gray}{-0.03}) \\
\bottomrule
\end{tabular}
\end{table}

\subsection{Transport geometry.}
Table~\ref{tab:jpta_transport_geometry} shows that transported anchors help only when combined with JPTA's reliability controls. Fixed text anchors remove semantic transport and perform well below Full JPTA, while direct image prototypes, unweighted Procrustes, and unconstrained linear maps often over-transport and hurt accuracy. Weighted Procrustes is the strongest geometry-only variant but still lags Full JPTA, especially in sparse batches. This supports the central design: JPTA does not simply replace CLIP text anchors with batch-derived prototypes, but uses transported semantic anchors with weighting, shrinkage, prior refinement, and adaptive control.

\begin{table}[htbp]
\centering
\scriptsize
\setlength{\tabcolsep}{3.6pt}
\caption{\textbf{Transport-geometry audit for JPTA.}
Results are regime-level means over FGVC-Aircraft, EuroSAT, and Oxford Flowers.
Accuracies are percentages; values in parentheses denote the change relative to Full JPTA.
The audit separates transport geometry from JPTA's weighting, shrinkage, prior refinement, and adaptive controller.
Fixed text anchors underperform Full JPTA, while direct image prototypes, unweighted Procrustes, and unconstrained maps often over-transport and degrade accuracy.
Full JPTA is strongest because transported semantic anchors are used through a weighted, shrunk, and prior-controlled update rather than as an ungated replacement for the text geometry.}
\label{tab:jpta_transport_geometry}
\begin{tabular}{lcccccc}
\toprule
Regime & Full JPTA & Fixed text & Image proto. & Unweighted Proc. & Weighted Proc. & Linear map \\
\midrule
Very low (1--4), $B{=}64$
& \textbf{64.19}
& 45.19 (\textcolor{red}{-19.00})
& 44.23 (\textcolor{red}{-19.96})
& 49.16 (\textcolor{red}{-15.03})
& 54.47 (\textcolor{red}{-9.72})
& 48.95 (\textcolor{red}{-15.24}) \\
High (25--50), $B{=}1000$
& \textbf{58.24}
& 48.20 (\textcolor{red}{-10.04})
& 54.82 (\textcolor{red}{-3.42})
& 53.66 (\textcolor{red}{-4.58})
& 56.02 (\textcolor{red}{-2.22})
& 55.14 (\textcolor{red}{-3.10}) \\
All classes
& \textbf{60.13}
& 50.15 (\textcolor{red}{-9.98})
& 58.43 (\textcolor{red}{-1.70})
& 58.64 (\textcolor{red}{-1.49})
& 58.86 (\textcolor{red}{-1.27})
& 58.42 (\textcolor{red}{-1.71}) \\
\midrule
Overall mean
& \textbf{60.85}
& 47.85 (\textcolor{red}{-13.01})
& 52.49 (\textcolor{red}{-8.36})
& 53.82 (\textcolor{red}{-7.03})
& 56.45 (\textcolor{red}{-4.40})
& 54.17 (\textcolor{red}{-6.68}) \\
\bottomrule
\end{tabular}
\end{table}

\subsection{Initialization Robustness and Negative Adaptation}
\label{app:initialization_robustness}
We test dependence on initial predictions using controlled affinity corruption and post-hoc initial-quality stratification. These complement natural domain shift in Appx.~\ref{app:source_target_transfer}. The studies measure average robustness and observed failures; neither supplies a guarantee of safe adaptation.

\paragraph{Controlled affinity corruption.}
For FGVC-Aircraft, EuroSAT, and ImageNet, we use 1,000 paired tasks per dataset. For nested fractions $\rho\in\{0.1,0.2,0.4\}$ of samples, we exchange the initial top-1 probability with the probability of the most semantically similar incorrect class. This intervenes on initialization, rather than adding image-space adversarial perturbations. Degradation is the paired within-method difference $\Delta_{\mathrm{noise}}(\rho)=\mathrm{Acc}(0)-\mathrm{Acc}(\rho)$.

\begin{center}
\begin{minipage}{\linewidth}
\centering
\captionsetup{hypcap=false}
\captionof{table}{\textbf{Degradation under controlled initialization noise.} Values are mean accuracy losses (pp) from each method's own clean run over the three datasets; lower is better. The final column reports the percentage of tasks where JPTA is strictly worse than the corresponding corrupted zero-shot initialization.}
\label{tab:jpta_initialization_noise}
\small
\setlength{\tabcolsep}{9pt}
\begin{tabular}{lcccc}
\toprule
Corrupted fraction & CLIP loss & StatA loss & JPTA loss & JPTA harm rate \\
\midrule
10\% & 3.60 & 2.20 & \textbf{0.22} & 15.4\% \\
20\% & 7.15 & 5.32 & \textbf{0.76} & 14.1\% \\
40\% & 14.81 & 14.51 & \textbf{4.51} & 15.0\% \\
\bottomrule
\end{tabular}
\end{minipage}
\end{center}
JPTA has the smallest degradation in all nine dataset--noise comparisons. At 40\% corruption, its mean loss is 10.30 pp smaller than CLIP's and 10.00 pp smaller than StatA's. On Aircraft, the corresponding losses are 2.69, 5.28, and 5.91 pp for JPTA, CLIP, and StatA, respectively. The approximately stable task-level harm rates indicate no monotonic increase under this intervention, but remain nonzero. Losses relative to each method's clean run measure sensitivity, not absolute rankings of corrupted accuracy. Batch-level soft evidence, neighborhood consistency, prior refinement, and evidence-dependent shrinkage provide a plausible explanation for the observed mitigation, rather than an adversarial-robustness guarantee.

\paragraph{Initial-quality stratification.}
We divide each dataset's 1,000 Very-Low tasks into quartiles of initial zero-shot accuracy, giving 11,000 tasks across 11 datasets. Q1 denotes the lowest-quality quartile and Q4 the highest. The ground-truth labels used to form these quartiles are available only for post-hoc diagnosis, never to JPTA at inference. Macro gains over zero-shot CLIP are +15.16, +15.67, +14.47, and +9.04 pp from Q1 to Q4. In Q1, macro accuracy increases from 44.18\% to 59.34\%, with positive mean gains on 10/11 datasets.

\begin{center}
\begin{minipage}{\linewidth}
\centering
\captionsetup{hypcap=false}
\captionof{table}{\textbf{Observed reliability control and failure boundary.} Q1/Q4 are dataset-specific initial-accuracy quartiles. Gains are macro accuracy changes (pp); harm is the strict task-level rate of underperforming zero-shot CLIP. Displacement is the reported weighted angular anchor diagnostic.}
\label{tab:jpta_initial_quality}
\small
\setlength{\tabcolsep}{8pt}
\begin{tabular}{lcccc}
\toprule
Initial quality & Gain over CLIP & Transport trust & Displacement & Harm rate \\
\midrule
Q1 (lowest) & +15.16 & 0.477 & $0.029^\circ$ & 22.38\% \\
Q4 (highest) & +9.04 & 0.619 & $0.041^\circ$ & 3.87\% \\
\bottomrule
\end{tabular}
\end{minipage}
\end{center}
Lower-quality initialization is accompanied by lower transport trust and smaller weighted anchor movement, consistent with more conservative adaptation. Nevertheless, harm increases substantially in Q1. The extreme FGVC-Aircraft Q1 stratum starts at only 5.01\% zero-shot accuracy and loses another 1.47 pp under JPTA. Thus, positive macro gains can coexist with harmful individual tasks or a negative dataset stratum. JPTA limits error amplification on average but requires minimally informative semantic features; trust and shrinkage mitigate risk without explicitly detecting or eliminating negative adaptation.

\section{Scope and Limitations}
\label{app:limitations}

JPTA targets closed-set transductive adaptation, where the label space is fixed and each unlabeled batch is informative enough to reveal local class structure. While the regime-adaptive trust mechanism limits drift on sparse or ambiguous batches, gains remain small when batches have very weak visual evidence, strong domain shift, or many visually similar classes. In these cases, JPTA stays close to the original CLIP anchor instead of aggressively adapting, improving stability but limiting correction.

The method also assumes that class names and text prototypes offer a meaningful semantic prior. If prompts are poorly specified, labels overlap semantically, or the test distribution includes unknown classes, the transported anchor may not resolve ambiguity. Extending JPTA to open-set, open-vocabulary, or growing label spaces is an important future direction.

Finally, JPTA uses cached frozen VLM features and no gradient-based adaptation. This yields efficient inference and avoids test-time fine-tuning, but representation failures of the frozen backbone cannot be fixed directly. Future work could pair batch-aware semantic anchoring with lightweight feature adaptation, better uncertainty estimation, or privacy-preserving batch statistics for sensitive deployments.

\paragraph{Missing labels and negative adaptation.}
The fixed candidate vocabulary must contain the true class. If a label is absent, the present closed-set assignment cannot output it; unknown detection and dynamic label expansion require additional mechanisms and are not evaluated here. JPTA could follow a candidate-retrieval stage, but that extension is not established by the present experiments. Appx.~\ref{app:initialization_robustness} additionally shows that conservative transport is not a guarantee of non-negative adaptation: nearly uninformative semantic evidence can still lead to accuracy loss. Conversely, stronger initial models reduce headroom without removing the gains observed for the tested OpenCLIP and SigLIP2 checkpoints (Appx.~\ref{app:modern_backbones}).

\paragraph{Broader impact.}
This work introduces a method for robust closed-set test-time adaptation of vision–language classifiers. Its main benefit is improved robustness when labels are unavailable and test batches have shifted class composition. Potential harms stem from wider use of stronger visual recognition, including unvalidated deployment in sensitive classification or surveillance contexts. We do not introduce new datasets, collect human-subject data, or deploy the method. Any practical use should include domain-specific validation, privacy-preserving handling of batch statistics, and checks for bias and open-set failures.

\clearpage

%% file: neurips_2026.bbl
\begin{thebibliography}{73}
\providecommand{\natexlab}[1]{#1}
\providecommand{\url}[1]{\texttt{#1}}
\expandafter\ifx\csname urlstyle\endcsname\relax
  \providecommand{\doi}[1]{doi: #1}\else
  \providecommand{\doi}{doi: \begingroup \urlstyle{rm}\Url}\fi

\bibitem[Abdul~Samadh et~al.(2023)Abdul~Samadh, Gani, Hussein, Khattak, Naseer, Shahbaz~Khan, and Khan]{abdul2023align}
Jameel Abdul~Samadh, Mohammad~Hanan Gani, Noor Hussein, Muhammad~Uzair Khattak, Muhammad~Muzammal Naseer, Fahad Shahbaz~Khan, and Salman~H Khan.
\newblock Align your prompts: Test-time prompting with distribution alignment for zero-shot generalization.
\newblock \emph{Advances in Neural Information Processing Systems}, 36:\penalty0 80396--80413, 2023.

\bibitem[Ahamed et~al.(2026)Ahamed, Miriya~Thanthrige, Rodrigo, and Khan]{ahamed2026atpt}
Shihab~Aaqil Ahamed, Udaya Sampath K.~Perera Miriya~Thanthrige, Ranga Rodrigo, and Muhammad~Haris Khan.
\newblock A-tpt: Angular diversity calibration properties for test-time prompt tuning of vision-language models.
\newblock In \emph{International Conference on Learning Representations}, 2026.

\bibitem[Baklouti et~al.(2026)Baklouti, Zanella, and Ayed]{baklouti2026limo}
Ghassen Baklouti, Maxime Zanella, and Ismail~Ben Ayed.
\newblock Language-aware information maximization for transductive few-shot clip.
\newblock \emph{Transactions on Machine Learning Research}, 2026.

\bibitem[Bossard et~al.(2014)Bossard, Guillaumin, and Van~Gool]{bossard2014food}
Lukas Bossard, Matthieu Guillaumin, and Luc Van~Gool.
\newblock Food-101--mining discriminative components with random forests.
\newblock In \emph{European conference on computer vision}, pages 446--461. Springer, 2014.

\bibitem[Boudiaf et~al.(2020)Boudiaf, Ziko, Rony, Dolz, Piantanida, and Ben~Ayed]{boudiaf2020information}
Malik Boudiaf, Imtiaz Ziko, J{\'e}r{\^o}me Rony, Jos{\'e} Dolz, Pablo Piantanida, and Ismail Ben~Ayed.
\newblock Information maximization for few-shot learning.
\newblock \emph{Advances in Neural Information Processing Systems}, 33:\penalty0 2445--2457, 2020.

\bibitem[Cimpoi et~al.(2014)Cimpoi, Maji, Kokkinos, Mohamed, and Vedaldi]{cimpoi2014describing}
Mircea Cimpoi, Subhransu Maji, Iasonas Kokkinos, Sammy Mohamed, and Andrea Vedaldi.
\newblock Describing textures in the wild.
\newblock In \emph{Proceedings of the IEEE conference on computer vision and pattern recognition}, pages 3606--3613, 2014.

\bibitem[Dai and Yang(2025)]{dai2025free}
Qiyuan Dai and Sibei Yang.
\newblock Free on the fly: Enhancing flexibility in test-time adaptation with online em.
\newblock In \emph{Proceedings of the Computer Vision and Pattern Recognition Conference}, pages 9538--9548, 2025.

\bibitem[Deng et~al.(2009)Deng, Dong, Socher, Li, Li, and Fei-Fei]{deng2009imagenet}
Jia Deng, Wei Dong, Richard Socher, Li-Jia Li, Kai Li, and Li~Fei-Fei.
\newblock Imagenet: A large-scale hierarchical image database.
\newblock In \emph{2009 IEEE conference on computer vision and pattern recognition}, pages 248--255. Ieee, 2009.

\bibitem[Farina et~al.(2025)Farina, Mancini, Iacca, and Ricci]{farina2025rethinking}
Matteo Farina, Massimiliano Mancini, Giovanni Iacca, and Elisa Ricci.
\newblock Rethinking few-shot adaptation of vision-language models in two stages.
\newblock In \emph{Proceedings of the IEEE/CVF Conference on Computer Vision and Pattern Recognition}, pages 29989--29998, 2025.

\bibitem[Fei-Fei et~al.(2004)Fei-Fei, Fergus, and Perona]{fei2004learning}
Li~Fei-Fei, Rob Fergus, and Pietro Perona.
\newblock Learning generative visual models from few training examples: An incremental bayesian approach tested on 101 object categories.
\newblock In \emph{2004 conference on computer vision and pattern recognition workshop}, pages 178--178. IEEE, 2004.

\bibitem[Feng et~al.(2023)Feng, Yu, Liu, Khan, and Zuo]{feng2023diverse}
Chun-Mei Feng, Kai Yu, Yong Liu, Salman Khan, and Wangmeng Zuo.
\newblock Diverse data augmentation with diffusions for effective test-time prompt tuning.
\newblock In \emph{Proceedings of the IEEE/CVF International Conference on Computer Vision}, pages 2704--2714, 2023.

\bibitem[Gadre et~al.(2023)Gadre, Ilharco, Fang, Hayase, Smyrnis, Nguyen, Marten, Wortsman, Ghosh, Zhang, Orgad, Entezari, Daras, Pratt, Ramanujan, Bitton, Marathe, Mussmann, Vencu, Cherti, Krishna, Koh, Saukh, Ratner, Song, Hajishirzi, Farhadi, Beaumont, Oh, Dimakis, Jitsev, Carmon, Shankar, and Schmidt]{gadre2023datacomp}
Samir~Yitzhak Gadre, Gabriel Ilharco, Alex Fang, Jonathan Hayase, Georgios Smyrnis, Thao Nguyen, Ryan Marten, Mitchell Wortsman, Dhruba Ghosh, Jieyu Zhang, Eyal Orgad, Rahim Entezari, Giannis Daras, Sarah Pratt, Vivek Ramanujan, Yonatan Bitton, Kalyani Marathe, Stephen Mussmann, Richard Vencu, Mehdi Cherti, Ranjay Krishna, Pang~Wei Koh, Olga Saukh, Alexander Ratner, Shuran Song, Hannaneh Hajishirzi, Ali Farhadi, Romain Beaumont, Sewoong Oh, Alex Dimakis, Jenia Jitsev, Yair Carmon, Vaishaal Shankar, and Ludwig Schmidt.
\newblock {DataComp}: In search of the next generation of multimodal datasets.
\newblock \emph{arXiv preprint arXiv:2304.14108}, 2023.

\bibitem[Gao et~al.(2024)Gao, Geng, Zhang, Ma, Fang, Zhang, Li, and Qiao]{gao2024clip-adapter}
Peng Gao, Shijie Geng, Renrui Zhang, Teli Ma, Rongyao Fang, Yongfeng Zhang, Hongsheng Li, and Yu~Qiao.
\newblock Clip-adapter: Better vision-language models with feature adapters.
\newblock \emph{International Journal of Computer Vision}, 132\penalty0 (2):\penalty0 581--595, 2024.

\bibitem[Helber et~al.(2019)Helber, Bischke, Dengel, and Borth]{helber2019eurosat}
Patrick Helber, Benjamin Bischke, Andreas Dengel, and Damian Borth.
\newblock Eurosat: A novel dataset and deep learning benchmark for land use and land cover classification.
\newblock \emph{IEEE Journal of Selected Topics in Applied Earth Observations and Remote Sensing}, 12\penalty0 (7):\penalty0 2217--2226, 2019.

\bibitem[Hendrycks et~al.(2021{\natexlab{a}})Hendrycks, Basart, Mu, Kadavath, Wang, Dorundo, Desai, Zhu, Parajuli, Guo, et~al.]{hendrycks2021manyImagenet-r}
Dan Hendrycks, Steven Basart, Norman Mu, Saurav Kadavath, Frank Wang, Evan Dorundo, Rahul Desai, Tyler Zhu, Samyak Parajuli, Mike Guo, et~al.
\newblock The many faces of robustness: A critical analysis of out-of-distribution generalization.
\newblock In \emph{Proceedings of the IEEE/CVF international conference on computer vision}, pages 8340--8349, 2021{\natexlab{a}}.

\bibitem[Hendrycks et~al.(2021{\natexlab{b}})Hendrycks, Zhao, Basart, Steinhardt, and Song]{hendrycks2021naturalImagenet-A}
Dan Hendrycks, Kevin Zhao, Steven Basart, Jacob Steinhardt, and Dawn Song.
\newblock Natural adversarial examples.
\newblock In \emph{Proceedings of the IEEE/CVF conference on computer vision and pattern recognition}, pages 15262--15271, 2021{\natexlab{b}}.

\bibitem[Hu et~al.(2021)Hu, Gripon, and Pateux]{hu2021leveraging}
Yuqing Hu, Vincent Gripon, and St{\'e}phane Pateux.
\newblock Leveraging the feature distribution in transfer-based few-shot learning.
\newblock In \emph{International conference on artificial neural networks}, pages 487--499. Springer, 2021.

\bibitem[Huang et~al.(2025)Huang, Jiang, Jiang, Li, Khan, and Wang]{huang2025cosmic}
Fanding Huang, Jingyan Jiang, Qinting Jiang, Hebei Li, Faisal~Nadeem Khan, and Zhi Wang.
\newblock Cosmic: Clique-oriented semantic multi-space integration for robust clip test-time adaptation.
\newblock In \emph{Proceedings of the Computer Vision and Pattern Recognition Conference}, pages 9772--9781, 2025.

\bibitem[Kalantidis et~al.(2024)Kalantidis, Tolias, et~al.]{kalantidis2024labelZLap}
Yannis Kalantidis, Giorgos Tolias, et~al.
\newblock Label propagation for zero-shot classification with vision-language models.
\newblock In \emph{Proceedings of the IEEE/CVF Conference on Computer Vision and Pattern Recognition}, pages 23209--23218, 2024.

\bibitem[Kang et~al.(2021)Kang, Kwon, Min, and Cho]{kang2021relational}
Dahyun Kang, Heeseung Kwon, Juhong Min, and Minsu Cho.
\newblock Relational embedding for few-shot classification.
\newblock In \emph{Proceedings of the IEEE/CVF international conference on computer vision}, pages 8822--8833, 2021.

\bibitem[Karmanov et~al.(2024)Karmanov, Guan, Lu, El~Saddik, and Xing]{karmanov2024efficient}
Adilbek Karmanov, Dayan Guan, Shijian Lu, Abdulmotaleb El~Saddik, and Eric Xing.
\newblock Efficient test-time adaptation of vision-language models.
\newblock In \emph{Proceedings of the IEEE/CVF Conference on Computer Vision and Pattern Recognition}, pages 14162--14171, 2024.

\bibitem[Khattak et~al.(2023{\natexlab{a}})Khattak, Rasheed, Maaz, Khan, and Khan]{khattak2023maple}
Muhammad~Uzair Khattak, Hanoona Rasheed, Muhammad Maaz, Salman Khan, and Fahad~Shahbaz Khan.
\newblock Maple: Multi-modal prompt learning.
\newblock In \emph{Proceedings of the IEEE/CVF conference on computer vision and pattern recognition}, pages 19113--19122, 2023{\natexlab{a}}.

\bibitem[Khattak et~al.(2023{\natexlab{b}})Khattak, Wasim, Naseer, Khan, Yang, and Khan]{khattak2023self}
Muhammad~Uzair Khattak, Syed~Talal Wasim, Muzammal Naseer, Salman Khan, Ming-Hsuan Yang, and Fahad~Shahbaz Khan.
\newblock Self-regulating prompts: Foundational model adaptation without forgetting.
\newblock In \emph{Proceedings of the IEEE/CVF international conference on computer vision}, pages 15190--15200, 2023{\natexlab{b}}.

\bibitem[Krause et~al.(2013)Krause, Stark, Deng, and Fei-Fei]{krause20133d}
Jonathan Krause, Michael Stark, Jia Deng, and Li~Fei-Fei.
\newblock 3d object representations for fine-grained categorization.
\newblock In \emph{Proceedings of the IEEE international conference on computer vision workshops}, pages 554--561, 2013.

\bibitem[Kravets et~al.(2025)Kravets, Chen, and Namboodiri]{kravets2025rethinking}
Alexey Kravets, Da~Chen, and Vinay~P Namboodiri.
\newblock Rethinking few shot clip benchmarks: A critical analysis in the inductive setting.
\newblock In \emph{Proceedings of the IEEE/CVF International Conference on Computer Vision}, pages 1902--1911, 2025.

\bibitem[Lafon et~al.(2025)Lafon, Vargas~Hakim, Rambour, Desrosiers, and Thome]{lafon2025cliptta}
Marc Lafon, Gustavo~Adolfo Vargas~Hakim, Cl{\'e}ment Rambour, Christian Desrosiers, and Nicolas Thome.
\newblock {CLIPTTA}: Robust contrastive vision-language test-time adaptation.
\newblock \emph{Advances in neural information processing systems}, 2025.

\bibitem[Lazarou et~al.(2021)Lazarou, Stathaki, and Avrithis]{lazarou2021iterative}
Michalis Lazarou, Tania Stathaki, and Yannis Avrithis.
\newblock Iterative label cleaning for transductive and semi-supervised few-shot learning.
\newblock In \emph{Proceedings of the IEEE/CVF International Conference on Computer Vision}, pages 8751--8760, 2021.

\bibitem[Li et~al.(2023)Li, Lian, Lu, Bai, Chen, and Wang]{li2023graphadapter}
Xin Li, Dongze Lian, Zhihe Lu, Jiawang Bai, Zhibo Chen, and Xinchao Wang.
\newblock Graphadapter: Tuning vision-language models with dual knowledge graph.
\newblock \emph{Advances in Neural Information Processing Systems}, 36:\penalty0 13448--13466, 2023.

\bibitem[Liang et~al.(2024)Liang, Hou, Hu, Chang, Shan, and Chen]{liang2024umfc}
Jiachen Liang, Ruibing Hou, Minyang Hu, Hong Chang, Shiguang Shan, and Xilin Chen.
\newblock {UMFC}: Unsupervised multi-domain feature calibration for vision-language models.
\newblock \emph{Advances in Neural Information Processing Systems}, 37:\penalty0 114072--114093, 2024.

\bibitem[Liu et~al.(2020)Liu, Song, and Qin]{liu2020prototype}
Jinlu Liu, Liang Song, and Yongqiang Qin.
\newblock Prototype rectification for few-shot learning.
\newblock In \emph{European conference on computer vision}, pages 741--756. Springer, 2020.

\bibitem[Maharana et~al.(2025)Maharana, Zhang, Karlinsky, Feris, and Guo]{maharana2025batclip}
Sarthak Maharana, Baoming Zhang, Leonid Karlinsky, Rogerio Feris, and Yunhui Guo.
\newblock Batclip: Bimodal online test-time adaptation for clip.
\newblock In \emph{Proceedings of the IEEE/CVF International Conference on Computer Vision}, pages 1569--1579, 2025.

\bibitem[Maji et~al.(2013)Maji, Rahtu, Kannala, Blaschko, and Vedaldi]{maji2013fineaircraft}
Subhransu Maji, Esa Rahtu, Juho Kannala, Matthew Blaschko, and Andrea Vedaldi.
\newblock Fine-grained visual classification of aircraft.
\newblock \emph{arXiv preprint arXiv:1306.5151}, 2013.

\bibitem[Martin et~al.(2024)Martin, Huang, Shakeri, Pesquet, and Ben~Ayed]{martin2024transductive}
S{\'e}gol{\`e}ne Martin, Yunshi Huang, Fereshteh Shakeri, Jean-Christophe Pesquet, and Ismail Ben~Ayed.
\newblock Transductive zero-shot and few-shot clip.
\newblock In \emph{Proceedings of the IEEE/CVF Conference on Computer Vision and Pattern Recognition}, pages 28816--28826, 2024.

\bibitem[Mohammad et~al.(2026)Mohammad, Behera, Pradhan, Kumar, and Ahmed]{Mohammad_2026_CVPR}
Mohammed Rahman Sherif~Khan Mohammad, Ardhendu Behera, Sandip Pradhan, Swagat Kumar, and Amr Ahmed.
\newblock Training-only heterogeneous image-patch-text graph supervision for advancing few-shot learning adapters.
\newblock In \emph{Proceedings of the IEEE/CVF Conference on Computer Vision and Pattern Recognition (CVPR)}, pages 19613--19622, June 2026.

\bibitem[Nilsback and Zisserman(2008)]{nilsback2008automatedflowers}
Maria-Elena Nilsback and Andrew Zisserman.
\newblock Automated flower classification over a large number of classes.
\newblock In \emph{2008 Sixth Indian conference on computer vision, graphics \& image processing}, pages 722--729. IEEE, 2008.

\bibitem[Osowiechi et~al.(2024)Osowiechi, Noori, Hakim, Yazdanpanah, Bahri, Cheraghalikhani, Dastani, Beizaee, Ayed, and Desrosiers]{osowiechi2024watt}
David Osowiechi, Mehrdad Noori, Gustavo~A Hakim, Moslem Yazdanpanah, Ali Bahri, Milad Cheraghalikhani, Sahar Dastani, Farzad Beizaee, Ismail~B Ayed, and Christian Desrosiers.
\newblock {WATT}: Weight average test-time adaptation of {CLIP}.
\newblock \emph{Advances in neural information processing systems}, 37:\penalty0 48015--48044, 2024.

\bibitem[Parkhi et~al.(2012)Parkhi, Vedaldi, Zisserman, and Jawahar]{parkhi2012cats}
Omkar~M Parkhi, Andrea Vedaldi, Andrew Zisserman, and CV~Jawahar.
\newblock Cats and dogs.
\newblock In \emph{2012 IEEE conference on computer vision and pattern recognition}, pages 3498--3505. IEEE, 2012.

\bibitem[Radford et~al.(2021)Radford, Kim, Hallacy, Ramesh, Goh, Agarwal, Sastry, Askell, Mishkin, Clark, et~al.]{radford2021learningclip}
Alec Radford, Jong~Wook Kim, Chris Hallacy, Aditya Ramesh, Gabriel Goh, Sandhini Agarwal, Girish Sastry, Amanda Askell, Pamela Mishkin, Jack Clark, et~al.
\newblock Learning transferable visual models from natural language supervision.
\newblock In \emph{International conference on machine learning}, pages 8748--8763. PmLR, 2021.

\bibitem[Recht et~al.(2019)Recht, Roelofs, Schmidt, and Shankar]{recht2019imagenetv2}
Benjamin Recht, Rebecca Roelofs, Ludwig Schmidt, and Vaishaal Shankar.
\newblock Do imagenet classifiers generalize to imagenet?
\newblock In \emph{International conference on machine learning}, pages 5389--5400. PMLR, 2019.

\bibitem[Rodr{\'\i}guez et~al.(2020)Rodr{\'\i}guez, Laradji, Drouin, and Lacoste]{rodriguez2020embedding}
Pau Rodr{\'\i}guez, Issam Laradji, Alexandre Drouin, and Alexandre Lacoste.
\newblock Embedding propagation: Smoother manifold for few-shot classification.
\newblock In \emph{European conference on computer vision}, pages 121--138. Springer, 2020.

\bibitem[Roy and Etemad(2024)]{roy2024coprompt}
Shuvendu Roy and Ali Etemad.
\newblock Consistency-guided prompt learning for vision-language models.
\newblock In \emph{The Twelfth International Conference on Learning Representations}, 2024.

\bibitem[Saha et~al.(2025)Saha, Lawrence, Van~Horn, and Maji]{saha2025generate}
Oindrila Saha, Logan Lawrence, Grant Van~Horn, and Subhransu Maji.
\newblock Generate, transduct, adapt: Iterative transduction with vlms.
\newblock In \emph{Proceedings of the IEEE/CVF International Conference on Computer Vision}, pages 1369--1379, 2025.

\bibitem[Satorras and Estrach(2018)]{satorras2018few}
Victor~Garcia Satorras and Joan~Bruna Estrach.
\newblock Few-shot learning with graph neural networks.
\newblock In \emph{International conference on learning representations}, 2018.

\bibitem[Sharifdeen et~al.(2025)Sharifdeen, Munir, Baliah, Khan, and Khan]{sharifdeen2025otpt}
Ashshak Sharifdeen, Muhammad~Akhtar Munir, Sanoojan Baliah, Salman Khan, and Muhammad~Haris Khan.
\newblock O-tpt: Orthogonality constraints for calibrating test-time prompt tuning in vision-language models.
\newblock In \emph{Proceedings of the IEEE/CVF Conference on Computer Vision and Pattern Recognition}, pages 19942--19951, 2025.

\bibitem[Sheng et~al.(2025)Sheng, Liang, Wang, and He]{sheng2025rtpt}
Lijun Sheng, Jian Liang, Zilei Wang, and Ran He.
\newblock R-tpt: Improving adversarial robustness of vision-language models through test-time prompt tuning.
\newblock In \emph{Proceedings of the IEEE/CVF Conference on Computer Vision and Pattern Recognition}, pages 29958--29967, 2025.

\bibitem[Shin et~al.(2024)Shin, Zhao, Cromp, Vishwakarma, and Sala]{shin2024otter}
Changho Shin, Jitian Zhao, Sonia Cromp, Harit Vishwakarma, and Frederic Sala.
\newblock Otter: effortless label distribution adaptation of zero-shot models.
\newblock \emph{Advances in Neural Information Processing Systems}, 37:\penalty0 44064--44101, 2024.

\bibitem[Shu et~al.(2022)Shu, Nie, Huang, Yu, Goldstein, Anandkumar, and Xiao]{shu2022test}
Manli Shu, Weili Nie, De-An Huang, Zhiding Yu, Tom Goldstein, Anima Anandkumar, and Chaowei Xiao.
\newblock Test-time prompt tuning for zero-shot generalization in vision-language models.
\newblock \emph{Advances in Neural Information Processing Systems}, 35:\penalty0 14274--14289, 2022.

\bibitem[Silva-Rodriguez et~al.(2024)Silva-Rodriguez, Hajimiri, Ben~Ayed, and Dolz]{silva2024closerCLAP}
Julio Silva-Rodriguez, Sina Hajimiri, Ismail Ben~Ayed, and Jose Dolz.
\newblock A closer look at the few-shot adaptation of large vision-language models.
\newblock In \emph{Proceedings of the IEEE/CVF Conference on Computer Vision and Pattern Recognition}, pages 23681--23690, 2024.

\bibitem[Song et~al.(2023)Song, Xue, Wang, Sun, Ge, Shan, et~al.]{song2023meta}
Lin Song, Ruoyi Xue, Hang Wang, Hongbin Sun, Yixiao Ge, Ying Shan, et~al.
\newblock Meta-adapter: An online few-shot learner for vision-language model.
\newblock \emph{Advances in Neural Information Processing Systems}, 36:\penalty0 55361--55374, 2023.

\bibitem[Soomro et~al.(2012)Soomro, Zamir, and Shah]{soomro2012ucf101}
Khurram Soomro, Amir~Roshan Zamir, and Mubarak Shah.
\newblock Ucf101: A dataset of 101 human actions classes from videos in the wild.
\newblock \emph{arXiv preprint arXiv:1212.0402}, 2012.

\bibitem[Sreenivas and Biswas(2025)]{sreenivas2025rosita}
Manogna Sreenivas and Soma Biswas.
\newblock Efficient open set single image test time adaptation of vision language models.
\newblock \emph{Transactions on Machine Learning Research (TMLR)}, 2025.

\bibitem[Tschannen et~al.(2025)Tschannen, Gritsenko, Wang, Naeem, Alabdulmohsin, Parthasarathy, Evans, Beyer, Xia, Mustafa, H{\'e}naff, Harmsen, Steiner, and Zhai]{tschannen2025siglip2}
Michael Tschannen, Alexey Gritsenko, Xiao Wang, Muhammad~Ferjad Naeem, Ibrahim Alabdulmohsin, Nikhil Parthasarathy, Talfan Evans, Lucas Beyer, Ye~Xia, Basil Mustafa, Olivier H{\'e}naff, Jeremiah Harmsen, Andreas Steiner, and Xiaohua Zhai.
\newblock {SigLIP 2}: Multilingual vision-language encoders with improved semantic understanding, localization, and dense features.
\newblock \emph{arXiv preprint arXiv:2502.14786}, 2025.

\bibitem[Veilleux et~al.(2021)Veilleux, Boudiaf, Piantanida, and Ben~Ayed]{veilleux2021realistic}
Olivier Veilleux, Malik Boudiaf, Pablo Piantanida, and Ismail Ben~Ayed.
\newblock Realistic evaluation of transductive few-shot learning.
\newblock \emph{Advances in Neural Information Processing Systems}, 34:\penalty0 9290--9302, 2021.

\bibitem[Wang et~al.(2020)Wang, Shelhamer, Liu, Olshausen, and Darrell]{wang2020tent}
Dequan Wang, Evan Shelhamer, Shaoteng Liu, Bruno Olshausen, and Trevor Darrell.
\newblock Tent: Fully test-time adaptation by entropy minimization.
\newblock \emph{arXiv preprint arXiv:2006.10726}, 2020.

\bibitem[Wang et~al.(2019)Wang, Ge, Lipton, and Xing]{wang2019learningImagenet-Sketch}
Haohan Wang, Songwei Ge, Zachary Lipton, and Eric~P Xing.
\newblock Learning robust global representations by penalizing local predictive power.
\newblock \emph{Advances in neural information processing systems}, 32, 2019.

\bibitem[Xiao et~al.(2010)Xiao, Hays, Ehinger, Oliva, and Torralba]{xiao2010sun}
Jianxiong Xiao, James Hays, Krista~A Ehinger, Aude Oliva, and Antonio Torralba.
\newblock Sun database: Large-scale scene recognition from abbey to zoo.
\newblock In \emph{2010 IEEE computer society conference on computer vision and pattern recognition}, pages 3485--3492. IEEE, 2010.

\bibitem[Yoon et~al.(2024)Yoon, Yoon, Tee, Hasegawa-Johnson, Li, and Yoo]{yoon2024ctpt}
Hee~Suk Yoon, Eunseop Yoon, Joshua Tian~Jin Tee, Mark Hasegawa-Johnson, Yingzhen Li, and Chang~D. Yoo.
\newblock C-tpt: Calibrated test-time prompt tuning for vision-language models via text feature dispersion.
\newblock In \emph{International Conference on Learning Representations}, 2024.

\bibitem[Yu et~al.(2023)Yu, Lu, Jin, Chen, and Wang]{yu2023task}
Tao Yu, Zhihe Lu, Xin Jin, Zhibo Chen, and Xinchao Wang.
\newblock Task residual for tuning vision-language models.
\newblock In \emph{Proceedings of the IEEE/CVF conference on computer vision and pattern recognition}, pages 10899--10909, 2023.

\bibitem[Zanella and Ben~Ayed(2024{\natexlab{a}})]{zanella2024lowclip-lora}
Maxime Zanella and Ismail Ben~Ayed.
\newblock Low-rank few-shot adaptation of vision-language models.
\newblock In \emph{Proceedings of the IEEE/CVF Conference on Computer Vision and Pattern Recognition Workshops}, pages 1593--1603, 2024{\natexlab{a}}.

\bibitem[Zanella and Ben~Ayed(2024{\natexlab{b}})]{zanella2024testMTA}
Maxime Zanella and Ismail Ben~Ayed.
\newblock On the test-time zero-shot generalization of vision-language models: Do we really need prompt learning?
\newblock In \emph{Proceedings of the IEEE/CVF Conference on Computer Vision and Pattern Recognition}, pages 23783--23793, 2024{\natexlab{b}}.

\bibitem[Zanella et~al.(2024)Zanella, G{\'e}rin, and Ben~Ayed]{zanella2024boosting}
Maxime Zanella, Beno{\^\i}t G{\'e}rin, and Ismail Ben~Ayed.
\newblock Boosting vision-language models with transduction.
\newblock \emph{Advances in Neural Information Processing Systems}, 37:\penalty0 62223--62256, 2024.

\bibitem[Zanella et~al.(2025)Zanella, Fuchs, De~Vleeschouwer, and Ben~Ayed]{zanella2025realistic}
Maxime Zanella, Cl{\'e}ment Fuchs, Christophe De~Vleeschouwer, and Ismail Ben~Ayed.
\newblock Realistic test-time adaptation of vision-language models.
\newblock In \emph{Proceedings of the IEEE/CVF Conference on Computer Vision and Pattern Recognition}, pages 25103--25112, 2025.

\bibitem[Zhang et~al.(2024{\natexlab{a}})Zhang, Stepputtis, Sycara, and Xie]{DPE}
Ce~Zhang, Simon Stepputtis, Katia Sycara, and Yaqi Xie.
\newblock Dual prototype evolving for test-time generalization of vision-language models.
\newblock \emph{Advances in Neural Information Processing Systems}, 37:\penalty0 32111--32136, 2024{\natexlab{a}}.

\bibitem[Zhang et~al.(2025)Zhang, Xu, Liu, Peng, and Zhou]{zhang2025scap}
Chenyu Zhang, Kunlun Xu, Zichen Liu, Yuxin Peng, and Jiahuan Zhou.
\newblock Scap: Transductive test-time adaptation via supportive clique-based attribute prompting.
\newblock In \emph{Proceedings of the IEEE/CVF Conference on Computer Vision and Pattern Recognition}, pages 30032--30041, 2025.

\bibitem[Zhang et~al.(2022)Zhang, Zhang, Fang, Gao, Li, Dai, Qiao, and Li]{zhang2022tip}
Renrui Zhang, Wei Zhang, Rongyao Fang, Peng Gao, Kunchang Li, Jifeng Dai, Yu~Qiao, and Hongsheng Li.
\newblock Tip-adapter: Training-free adaption of clip for few-shot classification.
\newblock In \emph{European conference on computer vision}, pages 493--510. Springer, 2022.

\bibitem[Zhang et~al.(2024{\natexlab{b}})Zhang, Wang, Guo, Dai, Chen, and Xia]{zhang2024boostadapter}
Taolin Zhang, Jinpeng Wang, Hang Guo, Tao Dai, Bin Chen, and Shu-Tao Xia.
\newblock Boostadapter: Improving vision-language test-time adaptation via regional bootstrapping.
\newblock \emph{Advances in Neural Information Processing Systems}, 37:\penalty0 67795--67825, 2024{\natexlab{b}}.

\bibitem[Zhang et~al.(2024{\natexlab{c}})Zhang, Zhu, Tang, Ma, Zhou, and Zhang]{zhang2024dual}
Yabin Zhang, Wenjie Zhu, Hui Tang, Zhiyuan Ma, Kaiyang Zhou, and Lei Zhang.
\newblock Dual memory networks: A versatile adaptation approach for vision-language models.
\newblock In \emph{Proceedings of the IEEE/CVF conference on computer vision and pattern recognition}, pages 28718--28728, 2024{\natexlab{c}}.

\bibitem[Zhao et~al.(2024)Zhao, Wang, Zhu, and Yang]{zhao2024clipreward}
Shuai Zhao, Xiaohan Wang, Linchao Zhu, and Yi~Yang.
\newblock Test-time adaptation with {CLIP} reward for zero-shot generalization in vision-language models.
\newblock In \emph{International Conference on Learning Representations (ICLR)}, 2024.

\bibitem[Zhou et~al.(2022{\natexlab{a}})Zhou, Yang, Loy, and Liu]{zhou2022conditionalCoCoOp}
Kaiyang Zhou, Jingkang Yang, Chen~Change Loy, and Ziwei Liu.
\newblock Conditional prompt learning for vision-language models.
\newblock In \emph{Proceedings of the IEEE/CVF conference on computer vision and pattern recognition}, pages 16816--16825, 2022{\natexlab{a}}.

\bibitem[Zhou et~al.(2022{\natexlab{b}})Zhou, Yang, Loy, and Liu]{zhou2022learningcoop}
Kaiyang Zhou, Jingkang Yang, Chen~Change Loy, and Ziwei Liu.
\newblock Learning to prompt for vision-language models.
\newblock \emph{International Journal of Computer Vision}, 130\penalty0 (9):\penalty0 2337--2348, 2022{\natexlab{b}}.

\bibitem[Zhou et~al.(2025)Zhou, Ye, Li, Li, Zhu, Deng, Liu, and Lei]{zhou2025bayesian}
Lihua Zhou, Mao Ye, Shuaifeng Li, Nianxin Li, Xiatian Zhu, Lei Deng, Hongbin Liu, and Zhen Lei.
\newblock Bayesian test-time adaptation for vision-language models.
\newblock In \emph{Proceedings of the Computer Vision and Pattern Recognition Conference}, pages 29999--30009, 2025.

\bibitem[Zhu and Koniusz(2023)]{zhu2023transductive}
Hao Zhu and Piotr Koniusz.
\newblock Transductive few-shot learning with prototype-based label propagation by iterative graph refinement.
\newblock In \emph{Proceedings of the IEEE/CVF conference on computer vision and pattern recognition}, pages 23996--24006, 2023.

\bibitem[Ziko et~al.(2020)Ziko, Dolz, Granger, and Ayed]{ziko2020laplacian}
Imtiaz Ziko, Jose Dolz, Eric Granger, and Ismail~Ben Ayed.
\newblock Laplacian regularized few-shot learning.
\newblock In \emph{International conference on machine learning}, pages 11660--11670. PMLR, 2020.

\end{thebibliography}
